\documentclass{article}

\usepackage{arxiv}

\usepackage[utf8]{inputenc} 
\usepackage[T1]{fontenc}    
\usepackage{hyperref}       
\usepackage{url}            
\usepackage{booktabs}       
\usepackage{amsfonts}       
\usepackage{nicefrac}       
\usepackage{microtype}      

\usepackage{amsmath}
\usepackage{algorithm}
\usepackage{algorithmic}
\usepackage{amssymb}
\usepackage{mathtools}
\usepackage{amsthm}
\usepackage{xcolor,colortbl} 
\usepackage{wrapfig}
\usepackage{natbib}

\title{Solving Reasoning Tasks with a Slot Transformer}

\author{%
  Ryan Faulkner \\
  Deepmind\\
  \texttt{rfaulk@google.com} \\
  \And
  Daniel Zoran \\
  Deepmind \\
  \texttt{danielzoran@deepmind.com} \\
}
\begin{document}

\maketitle

\begin{abstract}
The ability to carve the world into useful abstractions in order to reason about time and space is a crucial component of intelligence. In order to successfully perceive and act effectively using senses we must parse and compress large amounts of information for further downstream reasoning to take place, allowing increasingly complex concepts to emerge. If there is any hope to scale representation learning methods to work with real world scenes and temporal dynamics then there must be a way to learn accurate, concise, and composable abstractions across time.  We present the Slot Transformer, an architecture that leverages slot attention, transformers and iterative variational inference on video scene data to infer such representations. We evaluate the Slot Transformer on CLEVRER, Kinetics-600 and CATER datesets and demonstrate that the approach allows us to develop robust modeling and reasoning around complex behaviours as well as scores on these datasets that compare favourably to existing baselines.  Finally we evaluate the effectiveness of key components of the architecture, the model's representational capacity and its ability to predict from incomplete input.
\end{abstract}

\section{Introduction}
\label{intro}

Reasoning over time is an indispensable skill when navigating and interacting with a complex environment.  However, rationalizing about the world becomes an intractable problem if we are incapable of compressing it into to a reduced set of relevant abstractions.  For this reason relational reasoning and abstracting high-level concepts from complex scene data is a critical area of machine learning research.  Past approaches use relational reasoning and composable scene representation and have met success on static datasets \citep{santoro2017simple, santoro2018relational, greff2020multiobject, slotattn20, monet19} however, those gains must now be extended across time; this introduces a large set of new complexities in the form of temporal dynamics and scene physics.  Some recent work has utilized transformers \citep{perceiver2021, vaswani2017, ding2020objectbased} which provide a potential path toward understanding the type of approach that can succeed at solving this problem. 

The goal of the approach presented in this paper involves learning to output useful spatio-temporal representations of the input scene sequence which are hypothesized to be critical to solving downstream tasks for scene understanding and generalisation in domains containing complex visual scenes and temporal dynamics. The target domains for the evaluation of this type of model we have chosen involve scenes of synthetic and real world objects and behaviours requiring complex scene understanding and abstract reasoning via question and answer datasets.  This work combines a number of existing ideas in a novel way, namely slot attention \citep{slotattn20}, transformers \citep{vaswani2017} and iterative variational inference \citep{iai}; in order to better understand conceptual scene representation and how it can be achieved and embedded in more complex systems.  Several hypotheses drive the direction for this work. First, information about scenes may be more efficiently represented as independent components rather than in a monolithic representation. Second, ingesting spatio-temporal input in such a way that there is no bias to any step of the sequence is critical (see supplementary material for analysis) when forming representations about space and time over long sequences. Finally, processing information in an iterative fashion can provide the means to recursively recombine information in a way that is useful for reasoning tasks.

The main contributions of this work are: 1) present the Slot Transformer architecture for spatio-temporal inference and reasoning and a framework for learning how to encode useful representations, 2) evaluate this model against downstream tasks that require video understanding and reasoning capabilities in order to be solved, and finally 3) demonstrate the role the components of the overall approach play in the problem solving capabilities induced during training.  We have chosen three tasks on which to evaluate the Slot Transformer: CLEVRER \citep{yi2020clevrer}, a video dataset where questions are posed about objects in the scene, Kinetics-600 comprising of YouTube video data for action classification \citep{kinteics600} and CATER \citep{cater}, object relational data which requires video understanding of scene events to successfully solve.   In each of these cases we evaluate the Slot Transformer against current state-of-the-art approaches.

\section{Related Work}

We have drawn inspiration from models that induce scene understanding via object discovery.  In particular, the approaches taken by \citet{slotattn20, greff2020multiobject, monet19} all involve learning latent representations that allow the scene to be parsed into distinct objects.  Slot attention \citep{slotattn20} provides a means to extract relevant scene information into a set of latent vectors where each one queries scene pixels (or ResNet encoded \textit{super} pixels) and where the attention softmax is done across the query axis rather than the channel axis, inducing slots to compete to explain each pixel in the input. Another \emph{slotted} approach appears in IODINE \citep{greff2020multiobject}, which leverages iterative amortized inference \citep{iai, andrychowicz2016learning} to produce better posterior latent estimates for the slots which can then be used to reconstruct image by using a spatial Gaussian mixture using masks and means decoded from each slot as mixing weights and component means respectively.  We aim to extend this type of approach by applying the same ideas to sequential input, in particular, video input.

Self-attention and transformers \citep{vaswani2017, parisotto2019stabilizing} are also central to our work and have played a critical role in the field in recent years. Some of the latest applications have demonstrated the efficacy of this mechanism applied to problem solving in domains that require a capacity to reason as a success criteria \citep{santoro2018relational, clark2020transformers, russin2021compositional}. In particular, self-attention provides a means by which to form a pairwise relationship among any two elements in a sequence, however, this comes at a cost that scales quadratically in the sequence length, so care must be taken when choosing how to apply this technique.  One recent approach involving video sequence data, the TimeSformer from \citet{timesformer}, utilizes  attention over image patches across time applied successfully to sequence action classification \citep{goyal2017something, kinteics600}.

There has been a great deal of work on self-supervised video representation learning methods  \citep{wangLSTC, qianSCVRL, feichLSSUSRL, gopalSloTTAr} among the family of video representation learning models. In particular in \citet{qianSCVRL} the authors have explored using a contrastive loss and applied to Kinetics-600.  We include some of these results below in table \ref{table:kinetics} below. In \citet{kipfCOCLV} the authors present an architecure similar to ours, albeit without generative losses, where they show that their model achieves \textit{state-of-the-art} performance on image object segmentation in video sequences and optical flow field prediction.  In \citet{ding2020objectbased} the authors use an object representation model (\citet{monet19}) as input to a transformer and achieve very strong results on CLEVRER and CATER.

Neuro-symbolic logic based models \citep{garcez2020neurosymbolic} have been used to solve temporal reasoning problems and, in particular, to CLEVRER and CATER.  In section~\ref{section:ex} we examine the performance of the Dynamic Context Learner (DCL, \citet{chen2021grounding}) and Neuro-Symbolic Dynamic Reasoning (NS-DR, \citet{yi2020clevrer}) where NS-DR consists of neural networks for video parsing and dynamics prediction and a symbolic logic program executor and DCL is composed of Program Parser and Symbolic Executor while making use of extra labelled data when applied to CLEVRER.  Some components of these methods rely on explicit modeling of reasoning mechanisms crafted for the problem domain or on additional labelled annotations.  In contrast, the intent of our approach is to form an inductive bias around reasoning about sequences in general.

Recently \citet{perceiver2021} have published their work on the Perceiver, a model that makes use of attention asymmetry to ingest temporal multi-modal input.  In this work an attention bottleneck is introduced that reduces the dimensionality of the positional axis through which model inputs pass.  This provides tractability for inputs of large temporal and spatial dimensions where the quadratic scaling of transformers become otherwise prohibitive.  Notably the authors show that this approach succeeds on multimodal data, achieving state-of-art scores on the AudioSet dataset \citep{audioset} containing audio and video inputs.  Ours is a similar approach in spirit to the Perceiver in compressing representations via attention mechanisms however, we focus on the ability to reason using encoded sequences and primarily reduce over spatial axes.

\section{Model}

\begin{figure}
\centering
\includegraphics[width=1.\textwidth]{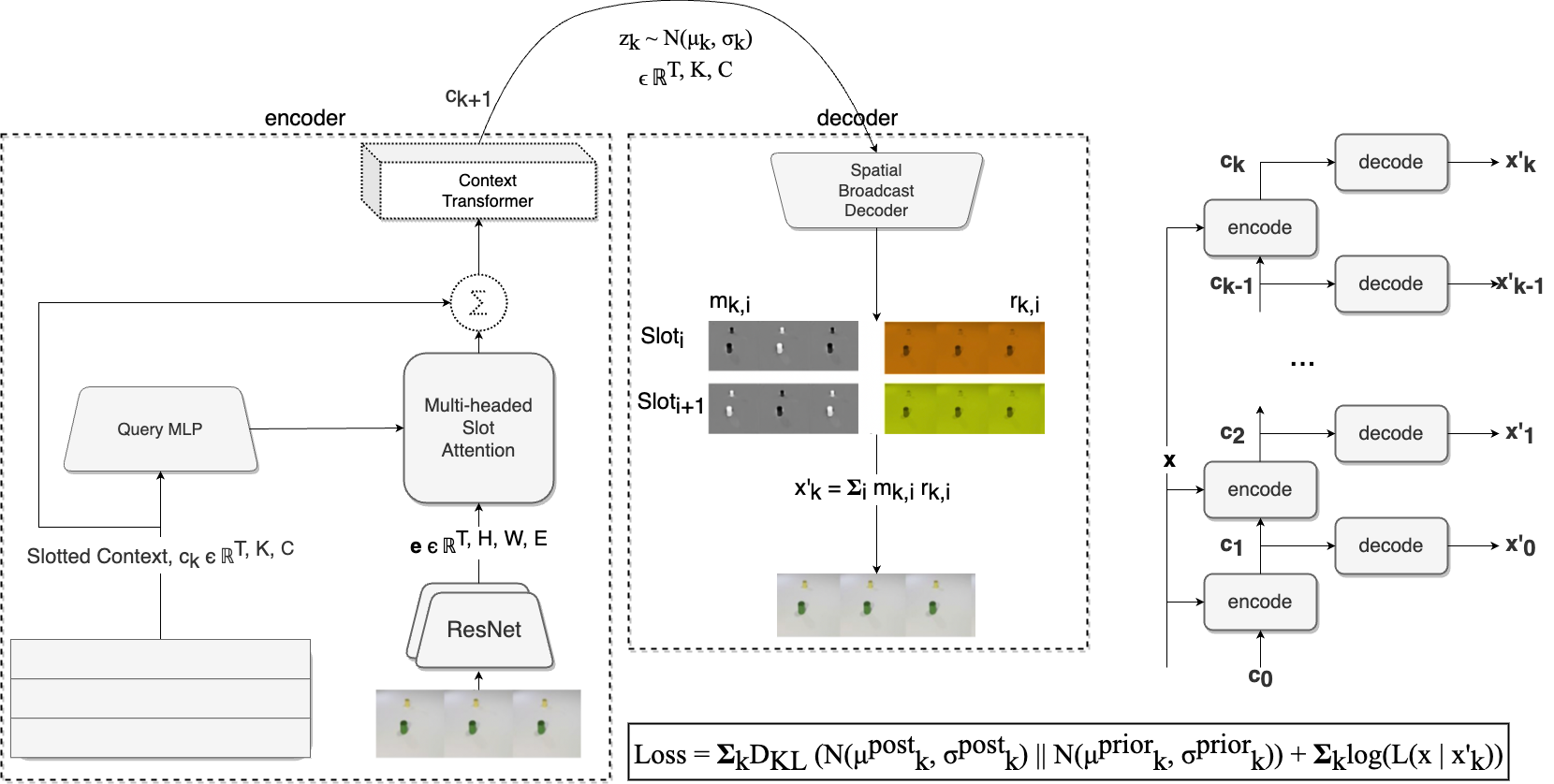}
\caption{The general model architecture with the encoding, decoding and iteration phases of the model.  The \textit{encode} phase computes an updated context $c_{k+1}$, the \textit{decode} phase produces a reconstruction $\mathbf{x}^{'}_{k}$.  Finally, the \textit{iterate} process repeats the encode-decode steps compressing more information into the context at each iteration.  KL and likelihood losses are computed across all iterations.}
\label{fig:architecture}
\end{figure}


We now present the Slot Transformer, a generative transformer model that leverages iterative inference to produce improved latent estimates given an input sequence. Our model can broadly be described in terms of three phases (see Figure \ref{fig:architecture}): an \textit{encode} phase where a spatio-temporal input is compressed to form a representation, then a \textit{decode} phase where the representation is used to reconstruct known data or predict unseen data; finally the \textit{iterate} phase, which encompasses the other two phases, whereby conditioning new representations on those from the previous iteration in the sequence input enable more useful representations to be inferred.  This process flow is similar in manner and inspired by the \textit{encode-process-decode} paradigm \citep{hamrickEPD}. We leverage the ideas of iterative attention used in slot attention and IODINE \citep{slotattn20, greff2020multiobject} based upon iterative variational inference to model good, compact representations for downstream tasks.

\subsection{Encoder}

The input sequence $\mathbf{X}\in \mathbb{R}^{T \times H \times W \times 3}$ is encoded once from raw pixels with a residual network \citep{resnet15} applied to each frame across the sequence: $\mathbf{e} = f_{\text{ResNet}}(\mathbf{X})$ where $\mathbf{e}\in \mathbb{R}^{T \times H' \times W' \times E}$ where $H'<H$ and $W'<W$.  To the resulting image encodings $\mathbf{e}$ we concatenate a spatial encoding basis comprised of Fourier basis functions dependent upon the spatial coordinates of the pixels.

Next we define the \textit{context}: a set of $T\times K$ vectors (or \emph{slots}), each of $C_{\text{context}}$ dimensions, that store contextual information about the scene. We denote the context as $\mathbf{c} \in \mathbb{R}^{T \times K \times C_{\text{context}}}$ where $T$ is the number of time steps in the input sequence. We will perform iterative updates on the context over the entire spatio-temporal volume with each forward pass through the model.  At each iteration, the encoder's role is to infer an updated spatio-temporal context via the context-transformer where the input is the combination of the context from the previous iteration plus the output of the slot attention on the input (more detail in \ref{section:iter}).

We initialize the context by sampling a standard Gaussian of size $K \times C_{\text{context}}$ then tiling this tensor over the same number of time steps as the input \footnote{It should be noted that while in this work the encoded representations retain the full time dimension this is not strictly necessary and this architecture may be modified to project to lower cardinality time dimensions in a straightforward manner.}. This yields the initial context $\mathbf{c}_0$. Sampling the context this way we break symmetry across the slots at each frame, and by tiling across time we encourage slots to have the same role across adjacent time-steps.  The slotted context provides an inductive bias to learn structured state information of the scene dynamics over time and space which the attention and context transform operations facilitate.

\subsubsection{Slot Attention}

Given the initial context, $\mathbf{c}_0$, we next apply a shared MLP across the slots to construct the queries, $\mathbf{q}_k$, one for each slot at each time step. We use these queries to attend over the keys and values decoded from the super pixels in $\mathbf{e}$ and their corresponding spatial position encodings.   The softmax step in this operation is applied across the slot dimension such that slots "compete" on which pixels in the scene they "explain" and losses that reward more efficient representations will induce the model to learn representations that best explain the scene \citep{slotattn20}.  One critical detail of this attention readout is that it occurs batched across time steps, learning temporal relationships is handled by the context transformer described below.  Much of this detail is captured in figure \ref{fig:architecture}.

The readout from the slot-attention step $\mathbf{a}_k$ for each slot at each time step is combined with the context via GRU style gating \citep{lstm97, gru14} with the gating function $\mathbf{c_{k}'}=f_{\text{gate}}(\mathbf{c}_{k}, \mathbf{a}_{k})$. 


\subsubsection{Context Transformer}

 Now that we have obtained the updated context $\mathbf{c'}$ using the input, it is passed through a transformer \citep{vaswani2017}, $\mathbf{c}_k = T_{\text{context}}(\mathbf{\mathbf{c'_k}})$, where positional encodings across the time sequence are applied but no masking.  This enables the model to form temporal connections across the sequence via the context slots.  At this step the same transformer weights are applied separately to each slot across time meaning that elements of each slot only communicates with other elements in that slot over the time axis.

The reason for this approach is two-fold, 1) this provides an inductive bias for the slots to adopt specialized roles when explaining the input, 2) this ensures that the complexity of the transform operation does not scale with the number of slots. At this step we could choose to introduce a temporal asymmetry as has shown promise in other work \citep{perceiver2021}.  This could yield better scalability and representational power if done correctly and we leave this as an avenue of future work.

One final note, the context transformer is only applied on the first iteration of the encoder as this allows the model to scale more easily to greater numbers of iterations, integrating information from the input through the slot attention operation alone on subsequent iterations.

\subsection{Decoder}
\label{section:decoder}

Once the encoder has generated an updated context we may hang new losses off of this representation by way of the decoder.  From here there are many possibilities for the way forward, for the scope of this this work we chose to explore image reconstruction over all frames in the sequence.  We hypothesize that these will help induce the model to learn useful and composable representations via the slots and to learn the temporal dynamics of the input via the context transformer.  We also decided to apply variational inference \citep{doersch2021tutorial} as we posit that this should help the model to better generalize by compressing the latent representations.  Therefore, we use the context to parameterize a Gaussian distribution (approximate posterior) by linearly projecting the context into parameters $\lambda_k \in \mathbb{R}^{2C_{\text{latent}}}$ and decode after sampling from it: $\mu_{k}, \log \sigma_{k} = \lambda_k = W_\lambda \mathbf{c}_{k}$ and $\mathbf{z}_{k} \sim N(\mu_{k}, \sigma_{k})$. While our choice of decoder is certainly dependent upon this choice we discuss this in more detail in the next section.

\subsubsection{Spatial Broadcast Decoder \& Masks}

Since the slots contain information that is likely spatially entangled we want to ensure that the encoder is left free to learn the spatio-temporal content based representations among the slots that can then be used by the decoder to rebuild the input.  For this reason we use a spatial broadcast decoder \citep{watters2019spatial} to distill information from the slots with the key property that the context channels are broadcast across spatial dimensions and decoded without upsampling - in effect each slot is allowed to explain independent parts of the image content.  The spatial decoder is batch applied across the batch, time and slot dimensions of the slotted latent to produce a one channel mask, $\mathbf{m}_{k,t}$, and an RGB mean image, $\mathbf{r}_{k,t}$: $\mathbf{m}_{k,t},\mathbf{r}_{k,t} = f_{\text{decoder}}(\mathbf{z}_{k, t})$.  These two elements are combined via weighted sum in a manner similar as is done in Slot Attention: $\mathbf{x'}_t = \sum_{k=1}^{K}\mathbf{m}_{k,t}  \mathbf{r}_{k,t}$.

\subsection{Iterative Model \& Losses}
\label{section:iter}

As mentioned in  Section~\ref{section:decoder} variational inference is applied to the model by projecting the context output from the encoder as Gaussian posterior parameter estimates over a latent distribution from which we sample $z_k$.  

We define a conditional prior, $\mathcal{P}(\mathbf{z}|\mathbf{x}_{1..p})$, by feeding the first $p < T$ steps of the context to the context transformer thereby limiting the prior to information within a subsequence of the input to form the \textit{prior context}.  To predict the remaining $T-p$ steps of the latent representation an auto-regressive model is used where the initial state is formed via a self attention operation on the prior context concatenated with a learnable token.  After the self attention operation is executed the output token is read and used as the initial state of the auto-regressive model. The output of this model is combined with the prior context to form the full definition of the latent prior parameters for the full sequence.


We make use of iterative inference \citep{iai} as applied by \citet{greff2020multiobject} where updates to the posterior parameters estimate, $\lambda_{k}$, are computed dependent on the sampled latents $z_{k}$, the input data $x$, the encoder architecture $f_{\phi}$, and auxiliary inputs $a_{k}$: 
\begin{equation}
\label{eqn:iai}
 \lambda_{k+1} = \lambda_{k} + f_{\phi}(z_{k}, x, a_{k}) \\
\end{equation}
On each iteration we also compute the reconstruction loss in addition to the Kullback-Liebler (KL) measure \citep{kl1951kl} of the posterior parameters with respect to the prior (i.e the ELBO):

\begin{equation}
\label{eqn:loss}
\mathcal{L}_\text{gen} = \sum_{t=1}^{T} \sum_{k=1}^{K}{\mathcal{D}_{KL}(\mathcal{N}(\mu^\text{post}_{k,t},\sigma^{\text{post}^2}_{k,t}) || \mathcal{N}(\mu^\text{prior}_{k,t},\sigma^{\text{prior}^2}_{k,t}))}
- \log{L(\mathbf{x}_t|\mathbf{x}_t')}
\end{equation}
We use a mixture distribution for the output likelihood $\log{L(\mathbf{x}_t|\mathbf{x}_t')}$, where the masks define a categorical distribution of components over Gaussian data distributions.


\subsection{Auxiliary Losses}

In addition to the variational and supervised losses we define auxiliary losses to help to induce better representations for scene understanding. For this we use the information from the slot specialization and language describing the scene when available.   

\subsubsection{Object Mask Prediction}

This loss is formed via model predictions on the latent slot values for selected target frames.  This is similar to the self-supervision strategies used in \citet{ding2020objectbased}.  For this we take the samples from the final iteration context, $\mathbf{z}_K$, then select $s < S$ slots at random masking out the last $k$ steps, for $k$ a fixed parameter; this is then fed once through the context transformer to compose $\mathbf{z}'_{K}$.  Next, from the masked steps we randomly select a subset of step indices, $\mathbf{T}_\text{targets}$, and compute the \textbf{L2} norm over the difference between the target predictions from $\mathbf{z}'_{K}$ and the true latents from $\mathbf{z}_{K}$:   $\mathcal{L}_\text{object} = \sum_{i}^{\mathbf{T}_\text{targets}} || \mathbf{z}_{K} - \mathbf{z}'_{K} ||$.  All of our top results were obtained when including this loss where predictions are done across the last half of the input sequence for each of $1$, $2$, and $3$ slots summed together.

\subsubsection{Question Prediction}

Given a Question-Answer dataset we hypothesize that we can learn better representations if they are good enough to predict the original question when conditioned on the answer.  During training, given the answer and the model latent $z_K$ we compose a new belief vector from these using a transformer and a CLS token.  This serves as the initial state of an auto-regressive model where we use teacher forcing from the question tokens and compute the sequence question word embeddings.  The cross-entropy between the predicted question embeddings and the true ones to form the question prediction loss: $\mathcal{L}_\text{question}$.

\subsection{Heads for Downstream Tasks}

\begin{figure}[h!]
\centering
\includegraphics[width=0.6\textwidth]{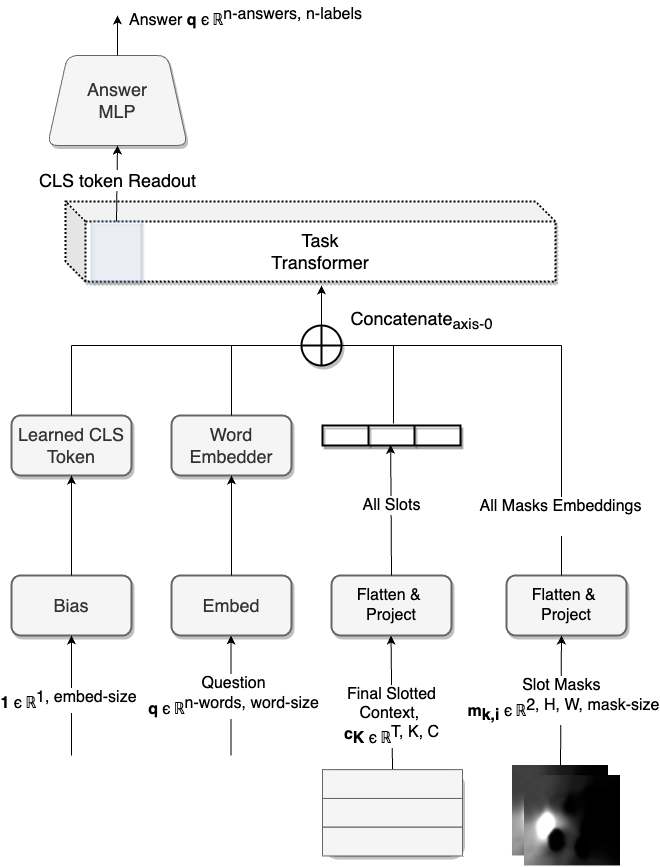}
\caption{General form of the \textit{task head} to compute answers given the output latent, $\textbf{z}_K$ from the encoder on the final iteration, an encoded question, the mask embeddings $\textbf{m}_K$ and a \textit{CLS} token.} 
\label{fig:heads}
\end{figure}

The context output from the final iteration, $c_{K}$, now forms an encoded sequence that may be used as input for specialized tasks.  The task head in figure \ref{fig:heads} depicts the general head architecture used for downstream tasks defined in Section~\ref{section:ex}.  The full input to the head consists of the question(s) provided with the task (when available), the final context from the encoder $\mathbf{c}_K$, the mask information from the last iteration of decoder output $\textbf{m}_K$, and a classification token, \textit{CLS}. As alluded to in Section~\ref{section:train}, we hypothesize at this point that $\mathbf{c}_K$ and $\mathbf{m}_K$ should contain information about the objects in the scene and the relations they share with eachother across the sequence.  Note that the time axis of $\mathbf{c}_k$ is now expanded to include all slots, so it is a sequence of length $T \times K$.  The token \textit{CLS} is prepended to the input sequence.

These elements are concatenated along the time axis, including absolute position encodings and now form the input to a multi-layer gated transformer \citep{parisotto2019stabilizing}. From the output the contents of the initial position corresponding to the \textit{CLS} token is then fed into an MLP (and softmaxed, depending on the task) to produce logits which can then be compared to the task labels via cross-entropy to produce a supervised loss $\mathcal{L}_\text{QA}$.

The generative, auxiliary, and supervised losses are then combined with scaling constants to yield the total loss:

\begin{equation}
\label{eqn:loss_tot}
\mathcal{L}_\text{total} = \lambda_\text{gen} \mathcal{L}_\text{gen} + \lambda_\text{question} \mathcal{L}_\text{question} + \lambda_\text{object} \mathcal{L}_\text{object} + \lambda_\text{QA} \mathcal{L}_\text{QA}
\end{equation}

Refer to 
the appendix for the dataset specific settings for the loss scaling constants.

\subsection{Training}
\label{section:train}

As discussed in section \ref{section:iter}, we define our prior distribution model by feeding only a proper subsequence of the input to the encoder while the posterior "sees" the full input sequence.  In practice the subsequence is chosen to be all but a final fixed length segment of the sequence.  Then sampling from the posterior distribution via the re-parameterization trick \citep{vaes2019kingma} we compute the log likelihood of the input sequence given the decoded masks and outputs from the sample.  Note that our final posterior estimates also include at least one iteration of approximate inference \citep{iai} and the full generative loss is composed of a sum over the iterations.  These loss terms are added to the supervised and auxiliary losses (eq. \ref{eqn:loss_tot}).  The KL term ensures that the conditional prior is capable of generating samples that accurately model the unknown portion of the sequence (figure \ref{fig:prior}). 
Thus the model must be able to encode information about the temporal dynamics of the scene objects \textit{as well as} the scene content in order to generalize well for scene understanding.

Samples from the posterior on the final model iteration are used as input to both the decoder and the task heads which also receive embeddings from the decoded masks.  For CLEVRER and CATER we use the LAMB optimizer \citep{lamb2019you} with weight decay to compute update gradients while stochastic gradient descent is applied to the Kinetics-600 dataset. The object prediction auxiliary loss is used over all datasets, while question prediction is only applicable to CLEVRER. We trained our models until they began to overfit which typically occured by at most $10e6$ model updates.  Detail on the experimental parameter values used can be found in the supplementary material.  



\begin{figure}
\centering
\includegraphics[width=1.\textwidth]{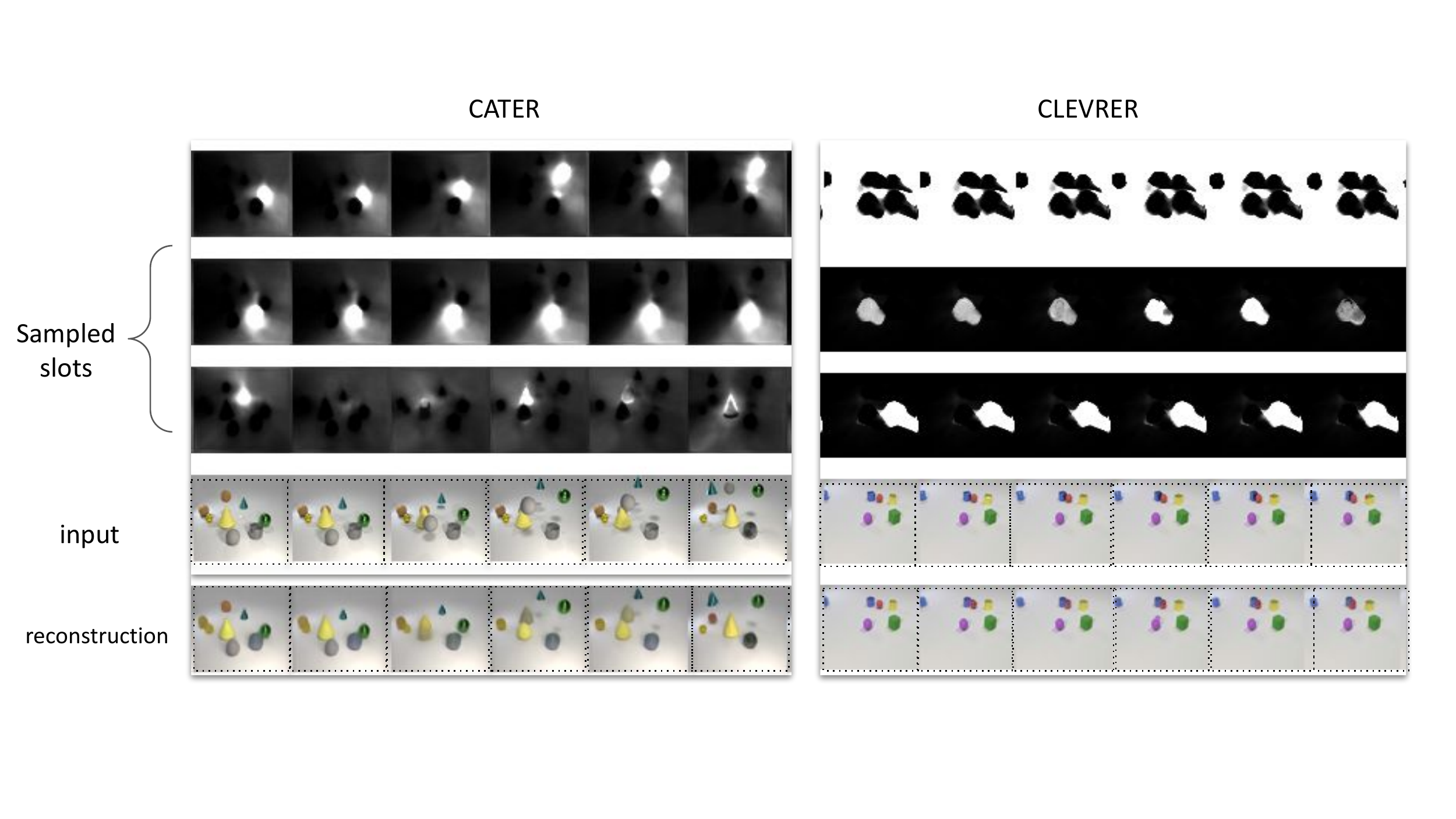}
\caption{Input, reconstructions, and decoder masks $\mathbf{m}$ of an eight slot model (only three shown for brevity). The slot masks partition the input space into content based components which combine to produce accurate reconstructions.}
\label{fig:masks}
\end{figure}

\section{Experiments}
\label{section:ex}

We trained and evaluated our Slot Transformer on three video datasets and tasks: 1) Kinetics-600: action classification on data from YouTube, CATER: object localization, CLEVRER: question and answer.  To solve these types of tasks requires cognitive abilities 
such as object identification, object permanence, and object relational reasoning across video sequences among others. 

\subsection{CLEVRER}

CLEVRER \citep{yi2020clevrer} is a visual question and answer video dataset that requires reasoning about the objects contained in the scene to answer a given question correctly. Much like its predecessor dataset, CLEVR \citep{johnson2016clevr}, CLEVRER consists of scenes containing objects of various shapes and colours that may move about a plain surface potentially resulting in collisions.  There are four types of questions posed: descriptive, explanatory, predictive and counterfactual, each challenging different powers of reasoning.  

We trained our model without any additional labelled data, simultaneously on all four CLEVRER question types with separate MLPs for multiple choice and descriptive question types.  We measure test performance and compare to the results from \citet{chen2021grounding} in Table \ref{table:clevrer-table}.  We computed results for correctness on the full question, that is, choosing the correct word answer for the descriptive task and selecting all the options correctly in the case of multiple choice.  



\begin{table}
  \caption{CLEVRER per Question Accuracies (\citet{chen2021grounding})}
  \label{table:clevrer-table}
  \centering
  \resizebox{\columnwidth}{!}{%
  \begin{tabular}{lllcccc}
    \toprule
    \rowcolor{lightgray} Methods & \multicolumn{2}{c}{Extra Labels} & Descriptive & Predictive & Explanatory & Counterfactual \\
    \rowcolor{lightgray} & Attr. & Prog. & & & & \\
    \midrule
    CNN+MLP & & & 48.4 & 18.3 & 13.2 & 9.0 \\
    CNN+LSTM & & & 51.8 & 17.5 & 31.6 & 14.7 \\
    Memory & No & No & 54.7 & 13.9 & 33.1 & 7.0 \\
    HCRN & & & 55.7 & 21.0 & 21.0 & 11.5 \\
    MAC (V) & & & 85.6 & 12.5 & 16.5 & 13.7 \\
    \rowcolor{yellow} \textbf{Slot Transformer (Ours)} & & & \bf{87.4} & \bf{48.3} & \bf{65.3} & \bf{45.2} \\
    \midrule
    TVQA+ & Yes & No & 72.0 & \bf{23.7} & \bf{48.9} & 4.1 \\
    MAC (V+) & & & \bf{86.4} & 22.3 & 42.9 & \bf{25.1} \\
    \midrule
    IEP (V) & & & 52.8 & 14.5 & 9.7 & 3.8 \\
    TbD-net (V) & No & Yes & 79.5 & 3.8 & 6.5 & 4.4 \\
    DCL & & & \bf{90.7} & \bf{82.8} & \bf{82.0} & \bf{46.5} \\
    \midrule
    NS-DR & & & 88.1 & 79.6 & 68.7 & 42.4 \\
    NS-DR (NE) & Yes & Yes & 85.8 & 74.3 & 54.1 & 42.0 \\
    DCL-Oracle & & & \bf{91.4} & \bf{82.0} & \bf{82.1} & \bf{46.9} \\
    \bottomrule
  \end{tabular}%
  }
\end{table}

\begin{figure}
\centering
\includegraphics[width=1\textwidth]{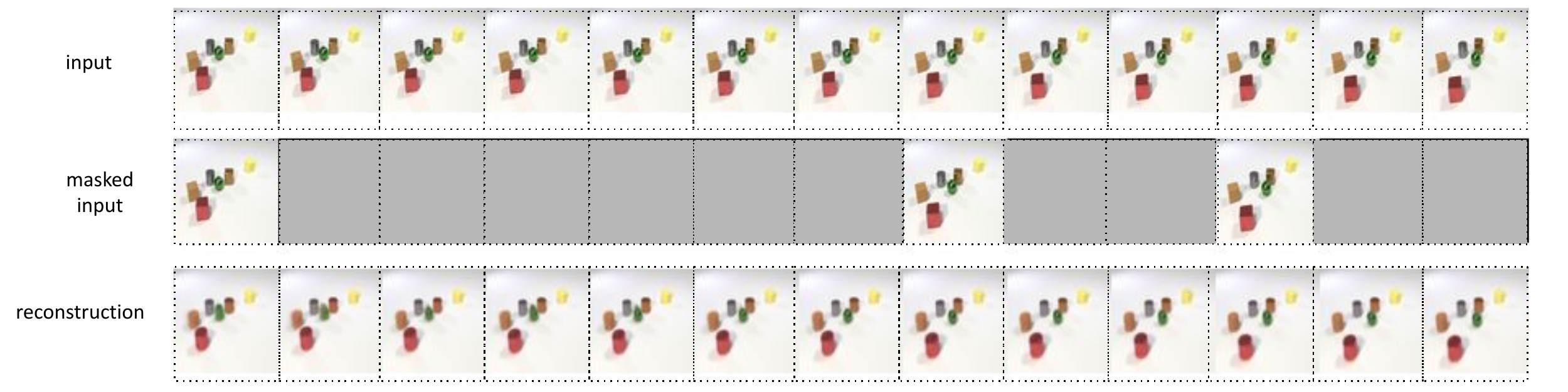}
\caption{\textbf{CLEVRER} eight slot model trajectory chunk.  These samples are drawn from a sequence unseen during training and the quality of the reconstructions remains qualitatively high.  Objects and significant contiguous chunks of frames are reconstructed accurately indicating that the model can infer the missing scene information.}
\label{fig:prediction}
\end{figure}

\subsection{CATER}

\begin{figure}
\centering
\includegraphics[width=1.\textwidth]{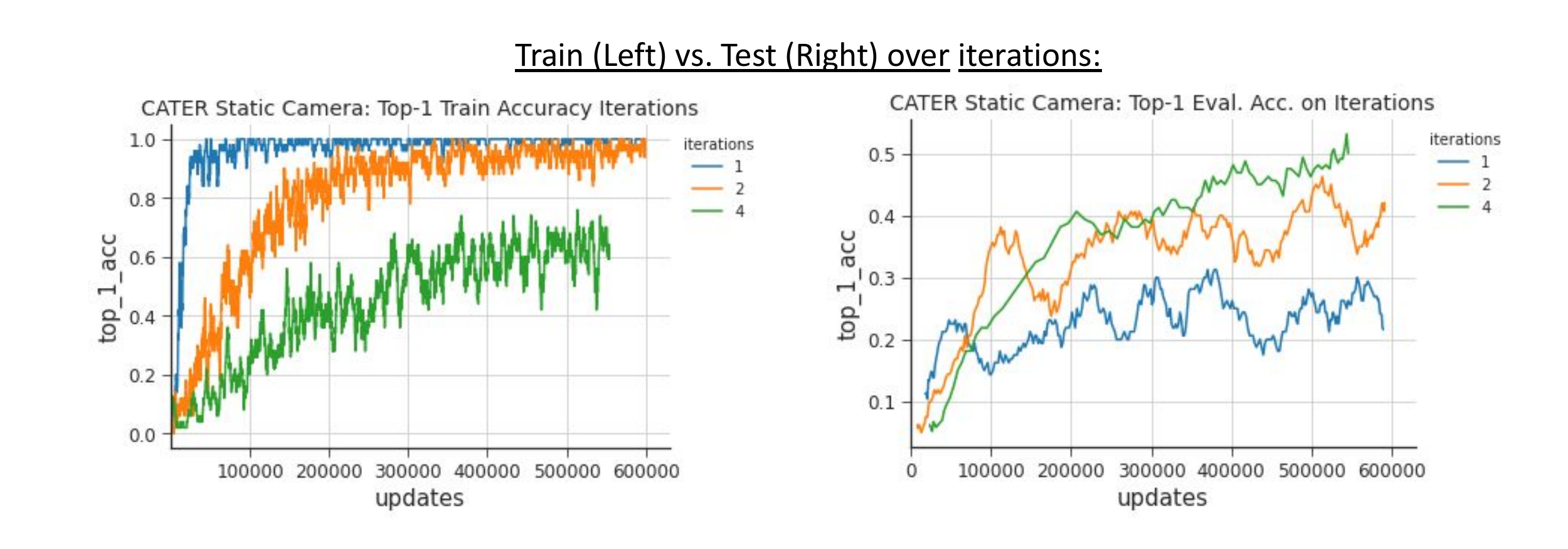}
\caption{\textbf{CATER} ten slot model Top-1 training and evaluation accuracy over $1,2$ and $4$ model iterations.  Models with fewer iterations tend to quickly overfit so while we see better performance on the training set the higher iteration models generalize significantly better and score higher on the unseen test set. }
\label{fig:iterations}
\end{figure}

\begin{table}
  \caption{CATER Results on Task 3: "Localization"}
  \label{table:cater-table}
  \centering
   \begin{tabular}{lllllll}
    \toprule
    \rowcolor{lightgray} Methods &  \multicolumn{3}{c}{\underline{Static Camera}} & \multicolumn{3}{c}{\underline{Moving Camera}} \\
    \rowcolor{lightgray} & Top-1 & Top-5 & L1 & Top-1 & Top-5 & L1 \\
    \midrule
    Random & 2.8 & 13.8 & 3.9 & - & - & - \\
    R3D LSTM & 60.2 & 81.8 & 1.2 & 28.6 & 63.3 & 1.7 \\
    R3D + NL LSTM & 46.2 & 69.9 & 1.5 & 38.6 & 70.2 & 1.5 \\
    Aloe \citet{ding2020objectbased} & \bf{74.0} & \bf{94.0} & \bf{0.44} & \bf{59.7} & \bf{90.1} & \bf{0.69} \\
    \midrule
    \rowcolor{yellow} \bf{Slot Transformer (Ours)} & 62.9 & 84.7 & 0.86 & 35.8 & 60.5 & 2.4 \\
    \bottomrule
  \end{tabular}
\end{table}

The Cater dateset (\citet{cater}) is composed of sequences of rendered objects, also based off of the earlier CLEVR dataset. The dataset is composed of three separate tasks: \textit{atomic action recognition}, \textit{compositional action recognition}, \textit{snitch localization}. In this paper we focus on the most challenging of the three, "snitch localization".  For this task, an object called the \textit{snitch} must be located on the final frame of the sequence on a $6 \times 6$ grid on the surface on which the objects move.  The grid is not actually visible in the image and the model must make it's best guess. The camera may be either moving or stationary.   


In the case of CATER the final MLP on the task head outputs $36$ labels over which we add a softmax layer (see Figure \ref{fig:heads}).  
Observing figure \ref{fig:iterations} it appears that in order to achieve good generalisation performance it is critical to apply multiple iterations.  
We compare our approach to existing baselines from the dataset paper \citep{cater} and we see that our results are competitive with existing state-of-the-art approaches. More details on the baselines and experimental setup can be found in the supplementary material.


\subsection{Kinetics-600}


Kinetics-600 \citep{kinteics600} is an action recognition dataset based on YouTube video.  The model is presented with a frame sequence and asked to classify the action in the scene.  Some previous approaches have used scaled up models or encoder pre-training setups \citep{kondratyuk2021movinets, arnab2021vivit} on larger YouTube or other video datasets prior to fine-tuning the classification head.  Our primary goal is to determine whether we can score in the range of stronger models using a similar setup to what we have applied on CLEVRER and CATER in the previous sections.  

We compare our results with those reported from \citet{kinteics600} on the Inflated 3D ConvNet \citep{i3d17}, the TimeSformer \citep{timesformer}, and baselines from \citet{qianSCVRL}.  Our results are approaching these baselines and it is noteworthy that the scores for these architectures have been pre-trained whereas ours have not.  We are intent on exploring what effect such a training regime may have on the slot transformer as future work.



\begin{figure}
\centering
\includegraphics[width=.75\textwidth]{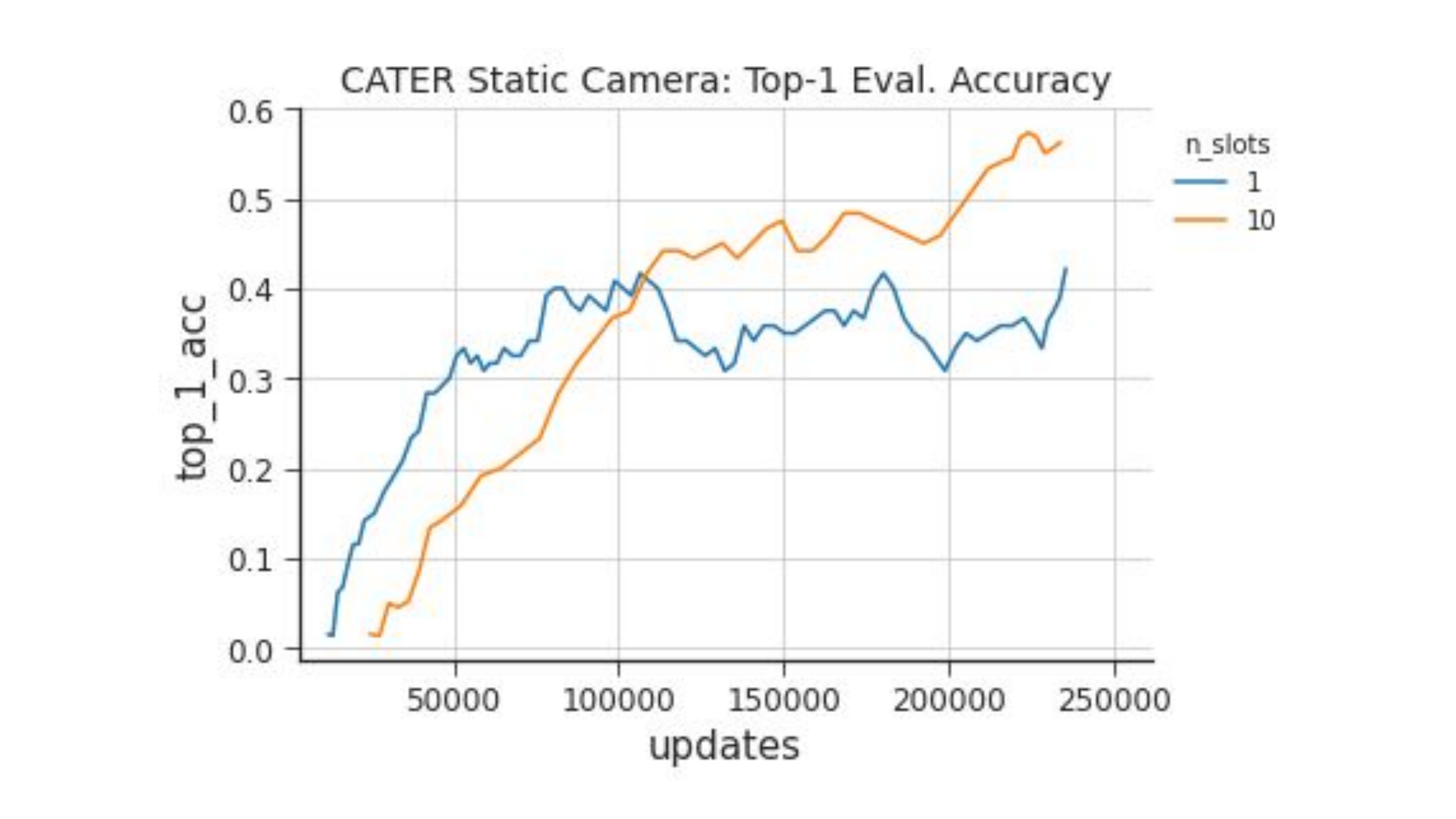}
\caption{\textbf{Ablate Attention}: Model accuracy when we ablate the slot attention, and return ResNet codes in place of the attention readout $\mathbf{y}_k$.}
\label{fig:ablate_attn}
\end{figure}

\section{Discussion}

\textbf{Scene and sequence representation.} Core to this work is the hypothesis that the combination of the context transformer and slot attention will 1) enable modeling of pairwise relations over the scene frames and the transmission of relevant information among the time-steps in the sequence and 2) provide a means to extract necessary object information from the scene to form a basis for scene understanding.  The slot transformer is able to learn good quality representations as evidenced by the model reconstructions and predictions (figure \ref{fig:masks}). Figure \ref{fig:masks} also suggests that the slot information partitions the scene in a sensible manner as different objects and content, such as shadows, are well defined in different slots as evidenced by the masks.  The predictions from the conditional prior in figure \ref{fig:prior}
(see appendix) demonstrate that the generative model functions as a good regularizer over scene data capable of encoding time sensitive scene data and using that information to infer missing data, at times over a large number of contiguous frames.

\begin{figure}
\centering
\includegraphics[width=0.73\textwidth]{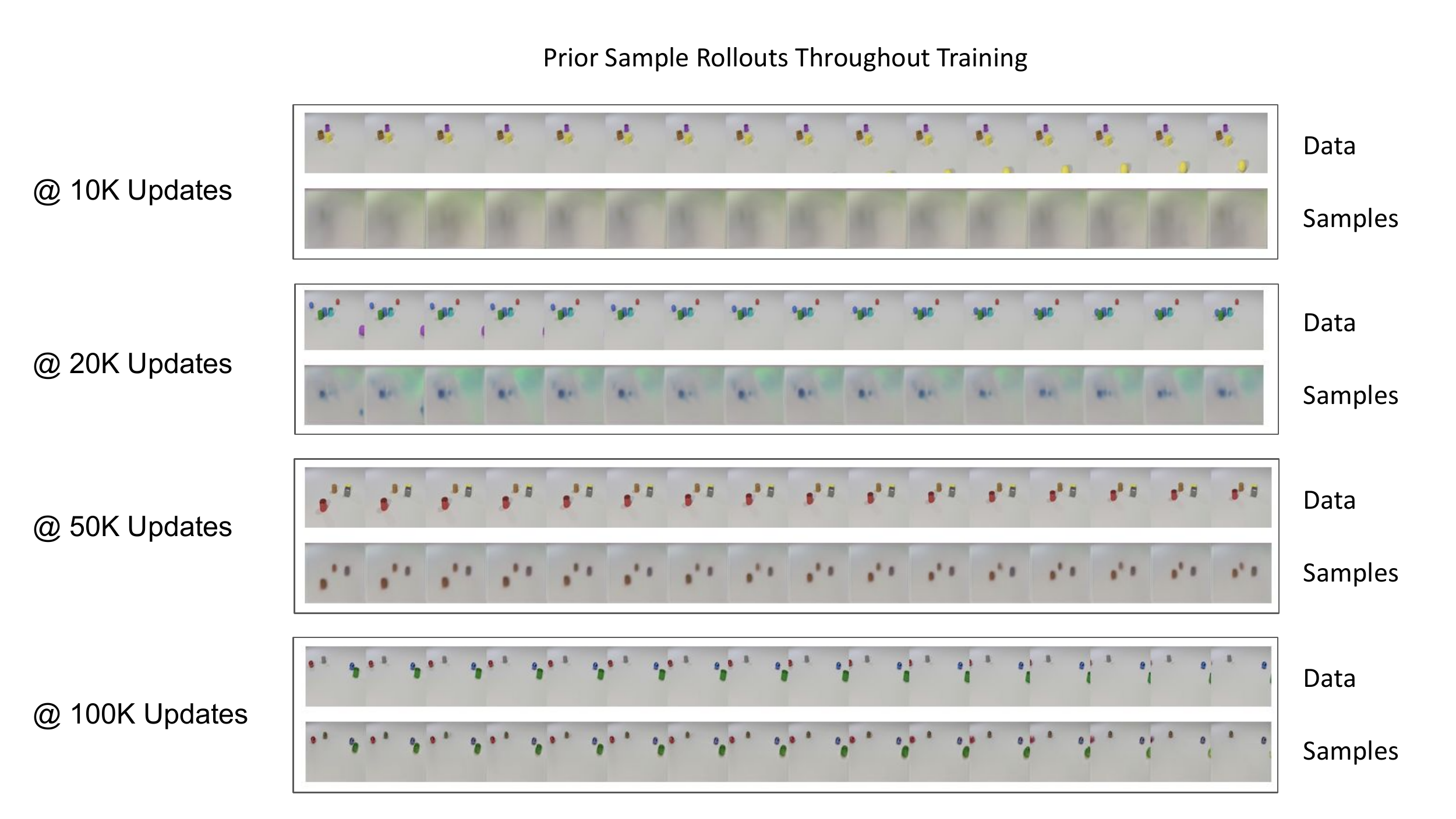}
\caption{\textbf{Prior Sampling}: Full 16 frame sampled rollouts from the test set drawn from the auto-regressive prior at various points in training on CLEVRER.  As the model learns its ability to predict unseen portions of the sequence given some information about the sequence improves markedly. }
\label{fig:prior}
\end{figure}

\textbf{Ablating the Slot Attention.} We ablate the attention by reducing to a single slot where the attention readout now simply returns the encoded frames from the ResNet.  This is compared to a ten slot model where the slot size has been normalized to ensure equal capacity for the comparison  (ie. the "slot size" of the one slot model is ten times as large) - the models are otherwise equivalent.  This ten slot model significantly outperforms the flat model by a margin of $57.5\%$ to $40.3\%$ (See the supplementary material for figure).

Figure \ref{fig:ablate_attn} shows that the slotted model clearly outperforms the flat model in evaluation on CATER as generalisation performance plateaus early for the non-attention model.



\textbf{Effect of more Iterations.} In the main text in section $5$ we noted that a ten slot model was trained on CATER-Static over $1$, $2$, and $4$ iterations with three seeds each.  The results are depicted in figure \ref{fig:iterations} for $1$, $2$, and $4$ iteration models.  When scoring the final model on the test set after training for the associated models we see scores of $51.2\% \pm 3.4$, $42.3\% \pm 2.9$, and $29.8\% \pm 2.1$ respectively.  This demonstrates a boost in model generalization when adding iterations to the model over training and evaluation.

\textbf{The Effect of the Iterative Model.} 
We ran a ten slot model on CATER-Static for $500K$ learner updates with batch size $32$ over $1$, $2$, and $4$ iterations with three seeds each.  After training with more iterations there is substantially less overfitting for the higher iteration models (figure in supplementary material).  The $1$, $2$, and $4$ iteration models scored $51.2\%$, $42.3\%$, and $29.8\%$ respectively.
We see in figure \ref{fig:iterations} that the iterative setup, in this case on CATER, helps generalisation performance on reasoning tasks - similar results were observed on CLEVRER.
Along with a boost in generalization performance additional iterations also appear to help stabilize performance likely due to the additional gradients and inference estimates available during training.

\textbf{Effect of the Generative Losses.} We sought to analyze the contribution of the generative losses to our model.  We did this ablation on CLEVRER-Descriptive by running the model in a deterministic mode where the output of the context isn't sampled but is instead passed on in a feed-forward manner.  In our ablation experiment we used a four iteration model averaged over three seeds for each condition.  Both models achieve near perfect QA accuracy on the training set where on the test set the generative model scores $55.8\%$ while the deterministic model scoring $43.8\%$. We hypothesize that generative losses help to provide a means to regularize over the latent space, enabling the model to learn better representations which are useful for reasoning tasks. 

\begin{table}
  \caption{Kinetic-600 Results. No pre-training was carried out on the Slot Transformer in contrast to the baselines.  We compare to models that benefit from pre-training: the TimeSformer \citep{timesformer}, the Inflated 3D ConvNet \citep{i3d17}, and from \citet{qianSCVRL} Contrastive Video Representation Learning (CVRL), SimCLR, and ImageNet models where the top model flavour is reported.}
  \label{table:kinetics}
  \centering
  \begin{tabular}{lll}
    \toprule
    \rowcolor{lightgray} Methods & Top-1 & Top-5  \\
    \midrule
    TimeSformer (+pre-train)  & \bf{82.2} & \bf{95.6} \\
    I3D (+pre-train)  & 71.7 & 90.4 \\
    \midrule
    SimCLR (+pre-train)  & 51.6 & - \\
    ImageNet (+pre-train) & 54.7 & - \\
    CVRL (+pre-train) & 72.9 & -\\
    \midrule
    \rowcolor{yellow} \bf{Slot Transformer (Ours)} & 68.2 & 87.6 \\
    \bottomrule
  \end{tabular}
\end{table}

\textbf{Effect of the Auxiliary Losses.} When including both object prediction and question prediction we achieved scores on the \textit{descriptive} CLEVRER task of $87.4\%$ compared with $45.4\%$ without any auxiliary losses, $53.7\%$ when including only question prediction, and $63.2\%$ when including only object prediction.  The improved performance on the test set is congruent with our hypothesis that these losses help to learn better latent representations for generalisation on reasoning tasks.


\textbf{Limitations.} It is evident that our results while strong are below \textit{state-of-the-art} on some tasks, a point which we'd like to address. As stated earlier, in all of our experiments we do not make use of any supplementary annotations, labels and nor do we do any pre-training.   Next, there are ways to better optimize the model: 1) scaling to more iterations, 2) further tuning the auxiliary losses, 3) optimizing better between generative, auxiliary and supervised losses, 4) pre-training the generative model - these are all areas of active experimentation along with applications to new domains.  We are encouraged by our success accurately modeling input data distributions, in particular in sequence prediction on sequences not seen in training, as is demonstrated in figures \ref{fig:masks} and \ref{fig:prior}
in both the main text and the supplementary material.


\textbf{Ethical Considerations.} While gaining an understanding and building systems to solve reasoning tasks is exciting work there are ethical implications to consider. We are categorically opposed to our work being used in the context of warfare or in fraud however, we understand that the possibility exists that a system which attempts to mimic reasoning could be used fraudulently.  While this is an inescapable risk when publishing work we hope that understanding these systems will outweigh the potential negative impact, and that by gaining deeper knowledge of it we can mitigate its misuse.


\textbf{Conclusion.} We have set out to demonstrate that we could learn to encode video scenes into compressed representations that store relevant spatial and temporal scene information and also that the biases induced by slot attention, transformers, and iterative inference would yield suitable representations for downstream processing in reasoning tasks with video input.  Taking this approach we have shown that we can achieve strong results on CATER and CLEVRER without using labelled data or specialized components and competitive results on Kinetics-600 with no model pre-training. We believe this work demonstrates promising first steps in this direction and we are eager to scale this effort.

\bibliography{main.bib}
\bibliographystyle{plainnat}

\appendix

\section{Model \& Experimental Details}

This section provides additional details and code for the model shown in figures 1 and 2 of the main text. 

\subsection{Features Encoder}

Given the input sequence, $\mathbf{x} \in \mathbb{R}^{T \times H \times W \times 3}$, we encode the input features using a $4$-Layer Residual Network (\citet{resnet15}):  $\mathbf{e} = f_{\text{ResNet}}(\mathbf{x})$.  The hyper-parameters per layer of this encoder are: output channels $ = (64, 128, 256, 512)$, residual blocks $=(2, 2, 2, 2)$,  and strides $=(2, 2, 2, 1)$.  We only do this once per input and we append a spatial Fourier basis (\cite{vaswani2017}) to the super-pixels in $\mathbf{e}$ via $f_{\text{spatial-encoding}}$ (see algorithm \ref{alg:enc-dec-iter}).

\subsection{Slot Attention}

First we batch apply the query MLP: $\mathbf{q_{k}} = f_{\text{MLP}}(\mathbf{c}_{k})$.  We use a $2$-layer MLP of sizes $(128, 128)$.  We apply queries independently across the sequence to their corresponding embeddings according to $t$.

For our attention readouts we take the approach of \citet{slotattn20}.  Namely, we perform the softmax operation over the attention weights then take the weighted mean of the weights before multiplying by the values:

\begin{align*}
    Q &= W_{Q} \mathbf{q}_{k} + \mathbf{b}_{Q} \\
    K &= W_{K} \mathbf{e} + \mathbf{b}_{K} \\
    V &= W_{V} \mathbf{e} + \mathbf{b}_{V} \\
    W^{\text{attn}} &= \text{softmax}(\frac{K Q^T}{\sqrt{s}}, \text{axis=slots}) \\
    W^{\text{attn}} &=   \frac{W^{\text{attn}}_{i,k}}{\sum_{l=1}^{I} W^{\text{attn}}_{l,k}} \\
    \mathbf{a}_k &= W_{\text{attn}}^T V \in \mathbb{R}^{T \times \text{slots} \times \text{s}} \\
\end{align*}

where $s$ is the slot size.  This is denoted as $f_{\text{SlotAttention}}$ in algorithm \ref{alg:enc-dec-iter} with input $\mathbf{q}_{k}$ and $\mathbf{e}$.

\subsection{Gating}

After computing the attention readout we use gating in a fashion similar to a Gated Recurrent Unit (\citet{gru14}).  Given the context $\mathbf{c}_{k}$, attention readout $\mathbf{a}_{k}$, and weights $W_{z,c}, W_{z,c}, W_{z,a}, W_{r,c}, W_{r,a}, W_{h,c}, W_{h,a}, \mathbf{b}_{z}, \mathbf{b}_{r}, \mathbf{b}_{h}$, we compute the updated context, $\mathbf{c}^{'}_{k}$ and do a final LayerNorm:

\begin{align*}
 \mathbf{z} &= \sigma(W_{z,c} \mathbf{c}_{k} + W_{z,a} \mathbf{a}_{k} + \mathbf{b}_{z}) \\
 \mathbf{r} &= \sigma(W_{r,c} \mathbf{c}_{k} + W_{r,a} \mathbf{a}_{k} + \mathbf{b}_{r}) \\
 \mathbf{h} &= \tanh{(W_{h,c} \mathbf{c}_{k} + W_{h,a} \mathbf{a}_{k} + \mathbf{b}_{h})} \\
 \mathbf{c}^{'}_{k} &= (\mathbf{1} - \mathbf{z}) \mathbf{c}_{k} + \mathbf{z} \mathbf{h} \\
 \mathbf{c}^{'}_{k} &= \text{LayerNorm}(\mathbf{c}^{'}_{k})  \in \mathbb{R}^{T \times \text{slots} \times \text{s}}\\
\end{align*}

Where $\sigma$ is the sigmoid function.  This is denoted as $f_{\text{gate}}$ in algorithm \ref{alg:enc-dec-iter} with input $\mathbf{q}_{k}$ and $\mathbf{e}$.

\subsection{Context Transformer}

When apply the context transformer to $\mathbf{c}^{'}_{k}$ we use a gated transformer (\citet{vaswani2017, parisotto2019stabilizing}) with absolute Fourier positional encodings concatenated to the input. The transformer is applied to each set of slots separately over the time dimension and the resulting sets are concatenated back together to form $\mathbf{c}_{k}^{''} = T_{\text{context}}(\mathbf{\mathbf{c'_k}})$ (see algorithm \ref{alg:enc-dec-iter}).  The transformer hyper-parameters per task are shown in appendix table \ref{table:hypers-table}.  Note that during training we use a dropout rate of $0.1$ on the attention weights \citet{vaswani2017}.

\subsection{Latent Sampling, Spatial Broadcast Decoding}

We first sample a latent from the posterior parameters derived from slotted context output from the context transformer, $\mathbf{c}_{k}^{''}$:

\begin{align*}
    \lambda_k &= W_\lambda \mathbf{c}_{k} \\
    \mathbf{\mu}_{k}, \log\mathbf{\sigma}_{k} &= \lambda_k \\
    \mathbf{z}_{k} &\sim N(\mathbf{\mu}_{k}, \mathbf{\sigma}_{k}) \in \mathbb{R}^{T \times \text{slots} \times \text{s}}\\
\end{align*}

To decode our latent we broadcast our latent to $\mathbb{R}^{T \times \text{slots} \times H \times W \times \text{s}}$, concatenate spatial Fourier encodings to all pixels, then batch apply a spatial decoder over time and slots (\citet{watters2019spatial}).  For the spatial broadcast decoder we use $4$ conv layers with the following parameters: output channels $=(64, 64, 64, 64)$, kernel shapes $=(3, 3, 3, 3)$, and strides $=(1, 1, 1, 1)$.

\subsection{Iterative Inference Update}

In the paper, in equation 1, we note that in addition to the auxiliary ELBO losses, $a_{k}$, we also compute an iterative inference update, an approximation of the loss on the prior with respect estimated latent posterior parameters, $\lambda_k$.  This loss estimate is dependent on the latent sample, $\mathbf{z}_k$, and the data $\mathbf{x}$. We have used the approach from \citet{iai} to compute an approximate update to our estimated latent posterior parameters and to derive the context for the next iteration, $\mathbf{c}_{k+1}$:

\begin{align*}
    \mathbf{\epsilon}_{x,k} &= \frac{\mathbf{x} - \mathbf{\mu}_{x}}{\mathbf{\sigma}_{x^2}} \\
    \mathbf{\epsilon}_{z,k} &= \frac{\mathbf{z} - \mathbf{\mu}_{z}}{\mathbf{\sigma}_{z^2}} \\
    \lambda_{k+1} &= \lambda_{k} + f_{\text{iterative}}(\mathbf{z}_{k}, \mathbf{x}_{k}, \mathbf{\epsilon}_{x,k},  \mathbf{\epsilon}_{z,k}) \\
    \mathbf{c}_{k+1} &= \text{LayerNorm}(\text{MLP}(\lambda_{k+1})) \in \mathbb{R}^{T \times \text{slots} \times \text{s}} \\
\end{align*}

Where is $f_{\text{iterative}}$ is a Linear network of size $\text{num-slots} \times \text{slot-size} (s)$.  Finally, note that all parameters are shared across iterations.

\subsection{Downstream Task Heads}

In figure 2 from the main text we show the general setup for the task head, more details can be found in algorithm \ref{alg:task-head}.  To form the input for the QA Transformer, $T_{\text{QA}}$, we project the slot values from the context, a classification token, and the first and last frame embeddings from the ResNet down to a common embedding size.   We also conditionally generate word embeddings for the input question as these only exist for CLEVRER tasks, these are otherwise omitted.  Once the $T_{\text{QA}}$ output has been computed, we extract its first frame corresponding to the token output $\in \mathbb{R}^{1\times \text{embed-size}}$, and feed that as input to the final MLP, $\text{MLP}_{\text{QA}}$, to produce logits against which we can compute a cross-entropy loss and accuracy. The hyper-parameters of the QA Transformer and the final MLP to generate the label logits is shown in table \ref{table:hypers-table}. Note that for the transformer during training we use a dropout rate of $0.1$ on the attention weights \citet{vaswani2017}.

\subsection{Losses \& Optimizer}

For CATER and CLEVRER we used the Lamb optimizer (\citet{you2020large}) with learning rates and weight decay values reported in table \ref{table:hypers-table}. All runs use an exponential decay rate for the first moment of past gradient of $0.9$, and for the second moment $0.999$ with an \textit{eps} constant of $1\text{e}^{-6}$.  For Kinetics-600 we used stochastic gradient descent with $8000$ burn-in steps to linearly ascend to the maximum learning rate and with a cosine decay schedule thereafter.

\subsection{Hardware Resources}

For our experiments we ran all learners on $4\times4$ DragonDonut TPUs with one full donut per learner.  We ran evaluators concurrently  on CPU, sampling a batch from the evaluation set every 30 seconds with evaluation batch sizes of $32$.

\subsection{Datset, Training \& Evaluation Details}

Tables \ref{table:tasks-table}, \ref{table:hypers-table}, and \ref{table:hypers-table-qa} describe the various hyper-parameter and dataset statistics relevant to our experiments.

\begin{table}
  \caption{Dataset \& task details.}
  \label{table:tasks-table}
  \centering
  \resizebox{\columnwidth}{!}{%
  \begin{tabular}{lccccc}
    \toprule
    Dataset & Sequence Length & Frame Size & \# Training Set Size & \# Evaluation Set Size & Label Size \\
    \midrule
    CLEVRER-Descriptive & $64$ & $64 \times 64$ & $10,000$ & $5,000$ & $42$ \\
    CLEVRER-Explanatory & $64$ & $64 \times 64$ & $10,000$ & $5,000$ & $1$ \\
    CLEVRER-Predictive & $64$ & $64 \times 64$ & $10,000$ & $5,000$ & $1$ \\
    CLEVRER-Counterfactual & $64$ & $64 \times 64$ & $10,000$ & $5,000$ & $1$ \\
    CATER Localization Task & $80$ & $64 \times 64$ & $3,850$ & $1,650$ & $36$ \\
    Kinetics-600 & $300$ & $256 \times 256$ & $392,622$ & $68,617$ & $600$ \\
    \bottomrule
  \end{tabular}%
  }
\end{table}

\begin{table}
  \caption{Hyper Parameters for the encode-decode-iterate model.}
  \label{table:hypers-table}
  \centering
  \resizebox{\columnwidth}{!}{%
  \begin{tabular}{lccccc|cccc|ccc}
    \toprule
     & & & & & & \multicolumn{4}{|c|}{Context Transformer} & & & \\
     \midrule
    Dataset & Learning Rate & Batch Size & Total Learner Steps & Weight Decay & Min. Mask Rate ($\rho$) & Layers & Heads & Embedding Size & MLP Layer Sizes & Slots & Slot Size & Iterations \\
    \midrule
    CLEVRER & $1e^{-4}$ & $8$ & $500K$ &$1e^{-4}$ & $1e^{-1}$ & $8$ & $4$ & $128$ & $(64, 128, 64)$ & $8$ & $64$ & $4$ \\
    CATER & $1e^{-4}$ & $8$ & $500K$ & $1e^{-2}$ & $5e^{-1}$ & $8$ & $4$ & $256$ & $(64, 256, 64)$ & $10$ & $64$ & $4$ \\
    Kinetics-600 & $1e^{-2}$ & $1M$ & $16M$ & $1e^{-2}$ & $5e^{-1}$ & $16$ & $4$ & $256$ & $(128, 256, 128)$ & $8$ & $128$ & $4$ \\
    \bottomrule
  \end{tabular}%
  }
\end{table}

\begin{table}
  \caption{Hyper Parameters for the task head.}
  \label{table:hypers-table-qa}
  \centering
  \resizebox{\columnwidth}{!}{%
  \begin{tabular}{lc|cccc|c}
    \toprule
     & & \multicolumn{4}{|c|}{QA Transformer} & \\
     \midrule
    Dataset & Input Embedding Size & Layers & Heads & Attention Embedding Size & MLP Layer Sizes & Final MLP Layer Sizes \\
    \midrule
    CLEVRER & $64$ & $8$ & $4$ & $128$ & $(128, 128)$ & $(2048, 2048, \text{label-size})$ \\
    CATER & $64$ & $8$ & $4$ & $256$ & $(256, 256)$ & $(2048, 2048, \text{label-size})$\\
    Kinetics-600 & $128$ & $16$ & $4$ & $256$ & $(256, 256)$ & $(2048, 2048, 2048, 2048 \text{label-size})$ \\
    \bottomrule
  \end{tabular}%
  }
\end{table}

\section{Code}
\label{section:code}

We now include pseudo-code that matches our implementation using the components described above.  See algorithms \ref{alg:enc-dec-iter} and \ref{alg:task-head}.

\begin{algorithm}
\caption{Pseudo code for the Encode-Decode-Iterate phases.}
\label{alg:enc-dec-iter}
\begin{algorithmic}[1]
  \STATE  \textbf{Input: }{$\mathbf{x}\in \mathbb{R}^{T \times \text{H} \times \text{W} \times 3}$}
  \STATE {$\mathbf{c}_{0} \sim N(0, 1) \in \mathbb{R}^{T \times K \times C_{\text{context}}}$}
  \STATE {$\mathbf{e} \gets f_{\text{ResNet}}(\mathbf{x})$}
  \STATE {$\mathbf{e} \gets \text{concat}([\mathbf{e}, f_{\text{spatial-encoding}}()], \text{axis=pixel})$}
  \FOR{k = 0 : K-1}
    \STATE {$\mathbf{q_{k}} \gets f_{\text{MLP}}(\mathbf{c}_{k})$} \COMMENT{Make Queries, batched over time}
    \STATE $\mathbf{a_{k}} \gets f_{\text{SlotAttention}}(\mathbf{q}_{k}, \mathbf{e})$ \COMMENT{$f_{\text{SlotAttention}}$ batched over time}
    \STATE $\mathbf{c_{k}'} \gets f_{\text{gate}}(\mathbf{c}_{k}, \mathbf{a}_{k})$
    \STATE $\mathbf{c}_{k}^{''} \gets T_{\text{context}}(\mathbf{\mathbf{c'_k}}, \text{position\_encode=True})$ \COMMENT{$T_{\text{context}}$ applied to each set of slots separately.}
    \STATE $\lambda_k \gets W_\lambda \mathbf{c}_{k}^{''}$ \COMMENT{Fetch posterior params and sample latent.}
    \STATE $\mu_{k}, \log\sigma_{k} \gets \lambda_k$
    \STATE $\mathbf{z}_{k} \sim N(\mu_{k}, \sigma_{k})$
    \STATE $\mathbf{m}_{k,t},\mathbf{r}_{k,t} \gets f_{\text{decoder}}(\mathbf{z}_{k, t})$ \COMMENT{Compute masks, assemble reconstructions.}
    \STATE $\mathbf{x}_t' \gets \sum_{k=1}^{K}\mathbf{m}_{k,t}\mathbf{r}_{k,t}$
    \STATE  $\lambda_{k+1} \gets \lambda_{k} + f_{\text{iterative}}(\mathbf{z}_{k}, \mathbf{x}_{k}, \mathbf{\epsilon}_{x,k},  \mathbf{\epsilon}_{z,k})$ \COMMENT{Iterative Inference Update}
    \STATE $\mathbf{c}_{k+1} \gets \text{LayerNorm}(\text{MLP}(\lambda_{k+1}))$ \COMMENT{New context}
  \ENDFOR
  \STATE $\mathcal{L}_{\text{ELBO}} = \sum_{t=1}^{T} \sum_{k=1}^{K}{\mathcal{D}_{KL}(\mathcal{N}(\mu_{k,t},\sigma_{k,t}^{2}) || \mathcal{N}(0,1)}  - \log{L(\mathbf{x}_t|\mathbf{x}_t')}$ 
\end{algorithmic}
\end{algorithm}

\begin{algorithm}
\caption{Downstream task heads.}
\label{alg:task-head}
\begin{algorithmic}[1]
  \STATE  \textbf{Input 1: }{$\mathbf{c}_{K} \in \mathbb{R}^{T \times \text{slots} \times \text{s}}$}
  \STATE  \textbf{Input 2: }{$\mathbf{e}_{\text{first}}\in \mathbb{R}^{2 \times H_{\text{ResNet}} \times W_{\text{ResNet}} \times 3}$}
  \STATE  \textbf{Input 3: }{$\mathbf{e}_{\text{last}} \in \mathbb{R}^{2 \times H_{\text{ResNet}} \times W_{\text{ResNet}} \times 3}$}
  \STATE  \textbf{Input 4: }{$\mathbf{question}\in \mathbb{R}^{\text{num-words} \times 1}$}
  \STATE  \textbf{Input 5: }{$\mathbf{answer}\in \mathbb{R}^{\text{label-size}}$}
  \STATE $\text{token}_{\text{in}} \gets \mathbf{1} \in \mathbb{R}^{1 \times \text{embed-size}}$
  \STATE $\text{token} \gets \text{Bias}(\text{token}_{in})$ \COMMENT{Use a learnable bias to shape the input token.}
  \IF{$\mathbf{question}$} \STATE {$\mathbf{question}^{\text{embed}} \gets f_{\text{embed}}(\mathbf{question}) \in \mathbb{R}^{\text{num-words} \times \text{embed-size}}$} \ELSE \STATE{$\mathbf{question}^{\text{embed}} \gets \emptyset$} \ENDIF
  \STATE $\mathbf{c_{\mathbf{K}}^{\text{embed}}} \gets \text{Reshape}(\text{Linear}_{\text{context}}(\mathbf{c}_{K})) \in \mathbb{R}^{(T \times \text{num-slots}) \times \text{embed-size}}$
  \STATE $\mathbf{e}_{\text{first}}^{\text{embed}} \gets \text{Linear}_{\text{frames}}(\text{Flatten}(\mathbf{e}_{\text{first}})) \in \mathbb{R}^{1 \times \text{embed-size}}$
  \STATE $\mathbf{e}_{\text{first}}^{\text{embed}} \gets \text{Linear}_{\text{frames}}(\text{Flatten}(\mathbf{e}_{\text{last}})) \in \mathbb{R}^{1 \times \text{embed-size}}$
  \STATE $\mathbf{x}_{\text{QA}} \gets \text{Concatenate}(\text{token}, \mathbf{question}^{\text{embed}}, \mathbf{c_{\mathbf{K}}^{\text{embed}}}, \mathbf{e}_{\text{first}}^{\text{embed}}, \mathbf{e}_{\text{last}}^{\text{embed}})$
  \STATE $\mathbf{y}_{\text{QA}} \gets \text{T}_{\text{QA}}(\mathbf{x}_{\text{QA}}, \text{position\_encode=True})$ \COMMENT{Run the QA Transformer.}
  \STATE $\text{logits} \gets \text{MLP}_{\text{QA}}(\mathbf{y}_{\text{QA}}[0, ...]) \in \mathbb{R}^{\text{label-size}} $  \COMMENT{Final MLP to produce logits.}
  \STATE $ \mathcal{L}_{\text{QA}} = \text{CrossEntropy}(\text{logits}, \text{labels})$ 
  \STATE $\mathcal{L}_{\text{QA}}$, logits
\end{algorithmic}
\end{algorithm}


\section{Additional Results \& Analysis}

We include performance curves below for train and evaluation on CLEVRER and Kinetics-600 in figures \ref{fig:desc_exp}, \ref{fig:pred_cfctl}, and \ref{fig:kinetics}.  We also include a figure illustrating the full set of masks over sequence snippets for CLEVRER and both CATER datasets in figure \ref{fig:all_masks}.

\subsection{Ablating the Context Transformer}

We carried out an additional analysis where we ablated the context transformer as we posited that this component would be critical in modeling temporal dynamics by allowing direct relations among any two frames in the sequence to be formed.  In order to measure its effect we replaced the Transformer with a Deep LSTM (\citet{lstm97}).  The Deep LSTM consists of an $4$-Layer model with hidden a state size of $256$, twice the size of the attention in the case of the transformer model.  The LSTM is rolled forward separately for each slot in a similar fashion to as is done for the context transformer and the slots are recombined to form the updated context.  Both were $2$ iteration models.

\begin{figure}[h!]
\centering
\includegraphics[width=1\textwidth]{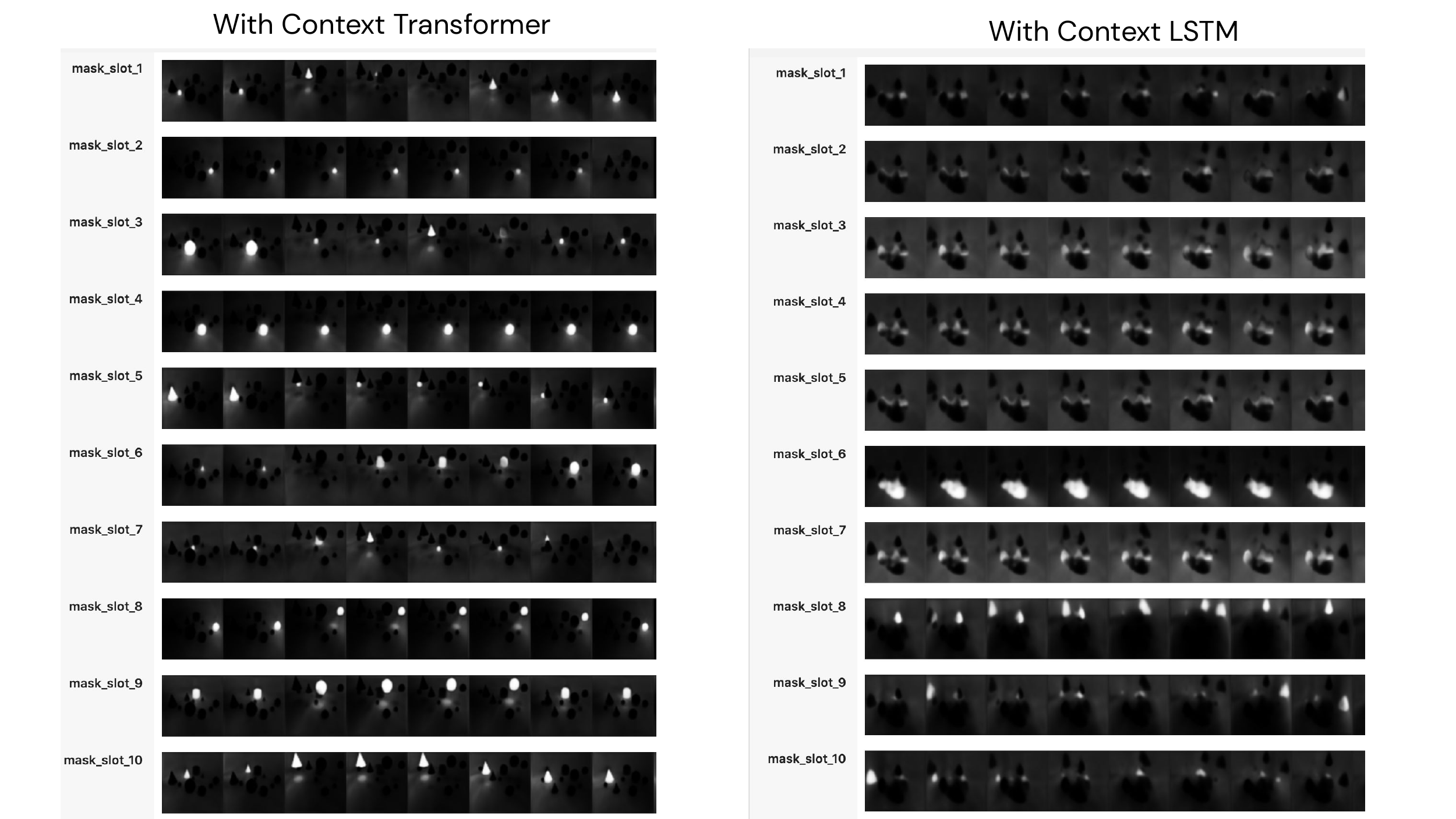}
\caption{Masks comparison when ablating the context Transformer for an LSTM.}
\label{fig:lstm_masks}
\end{figure}

\begin{figure}[h!]
\centering
\includegraphics[width=1\textwidth]{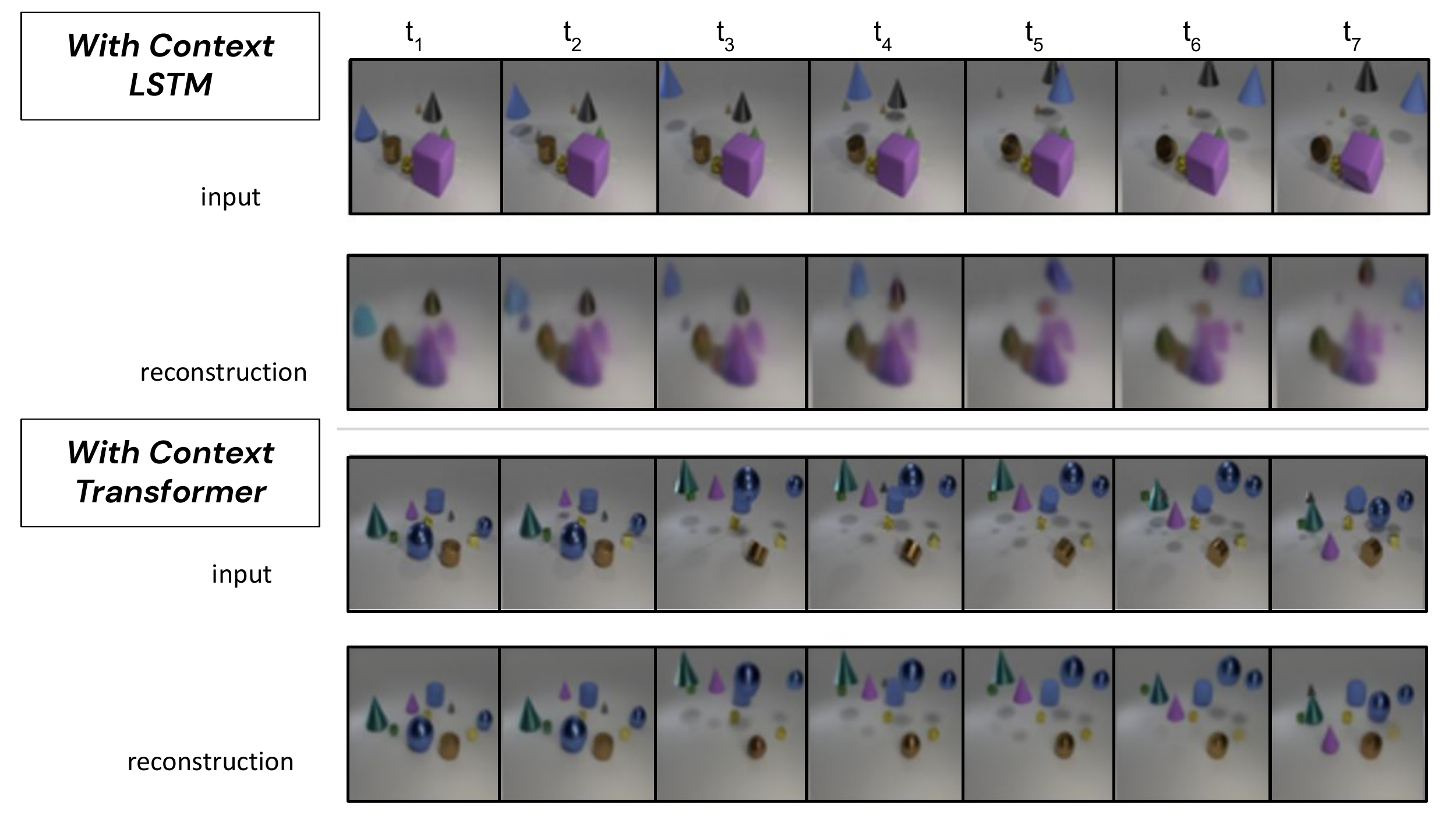}
\caption{Reconstructions comparison when ablating the context Transformer for an LSTM.}
\label{fig:lstm_recon}
\end{figure}

\begin{figure}[h!]
\centering
\includegraphics[width=1\textwidth]{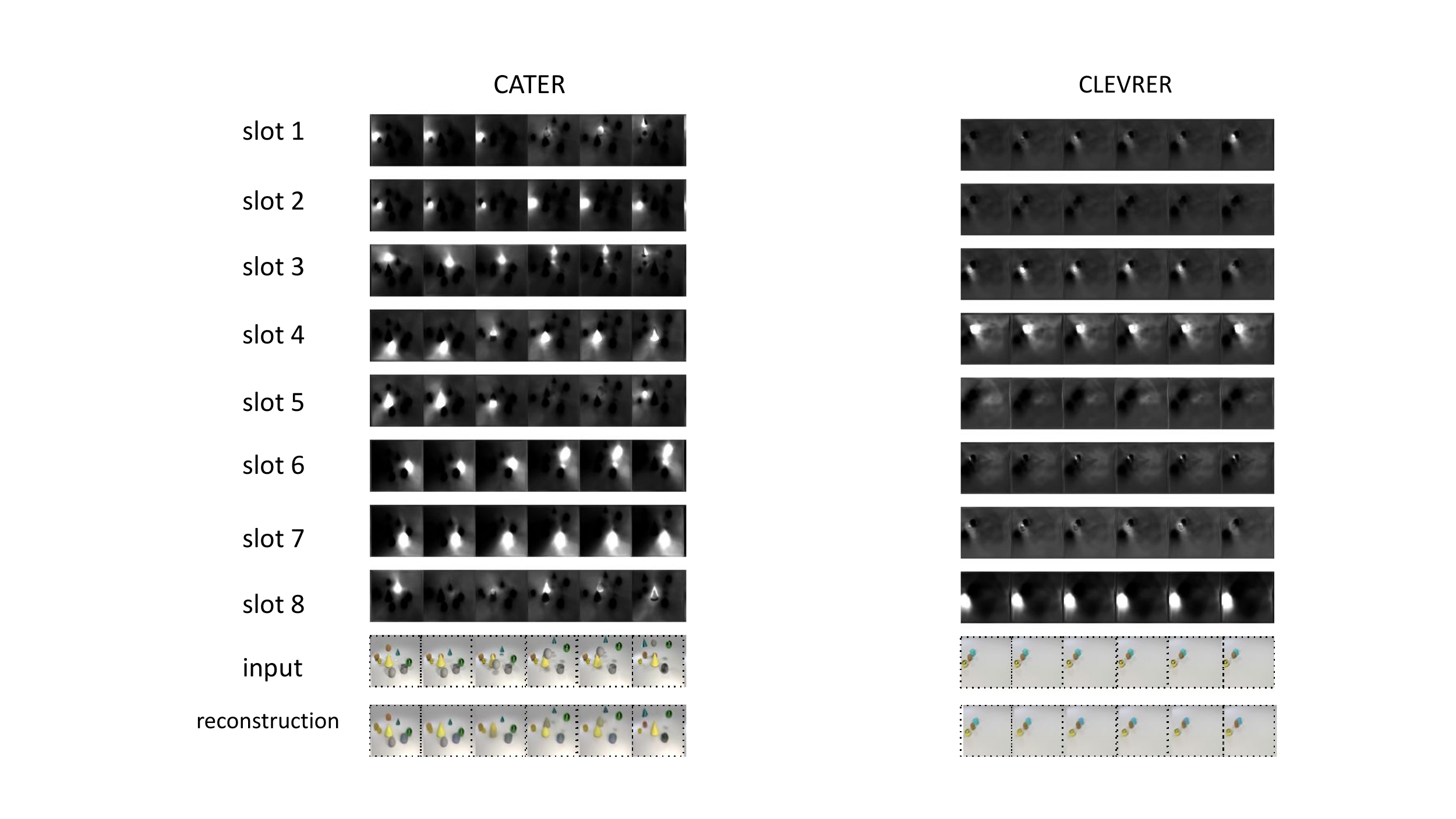}
\caption{All masks for $8$ slot models on CLEVRER \& CATER tasks with static and moving cameras.}
\label{fig:all_masks}
\end{figure}

\begin{figure}[h!]
\centering
\includegraphics[width=1\textwidth]{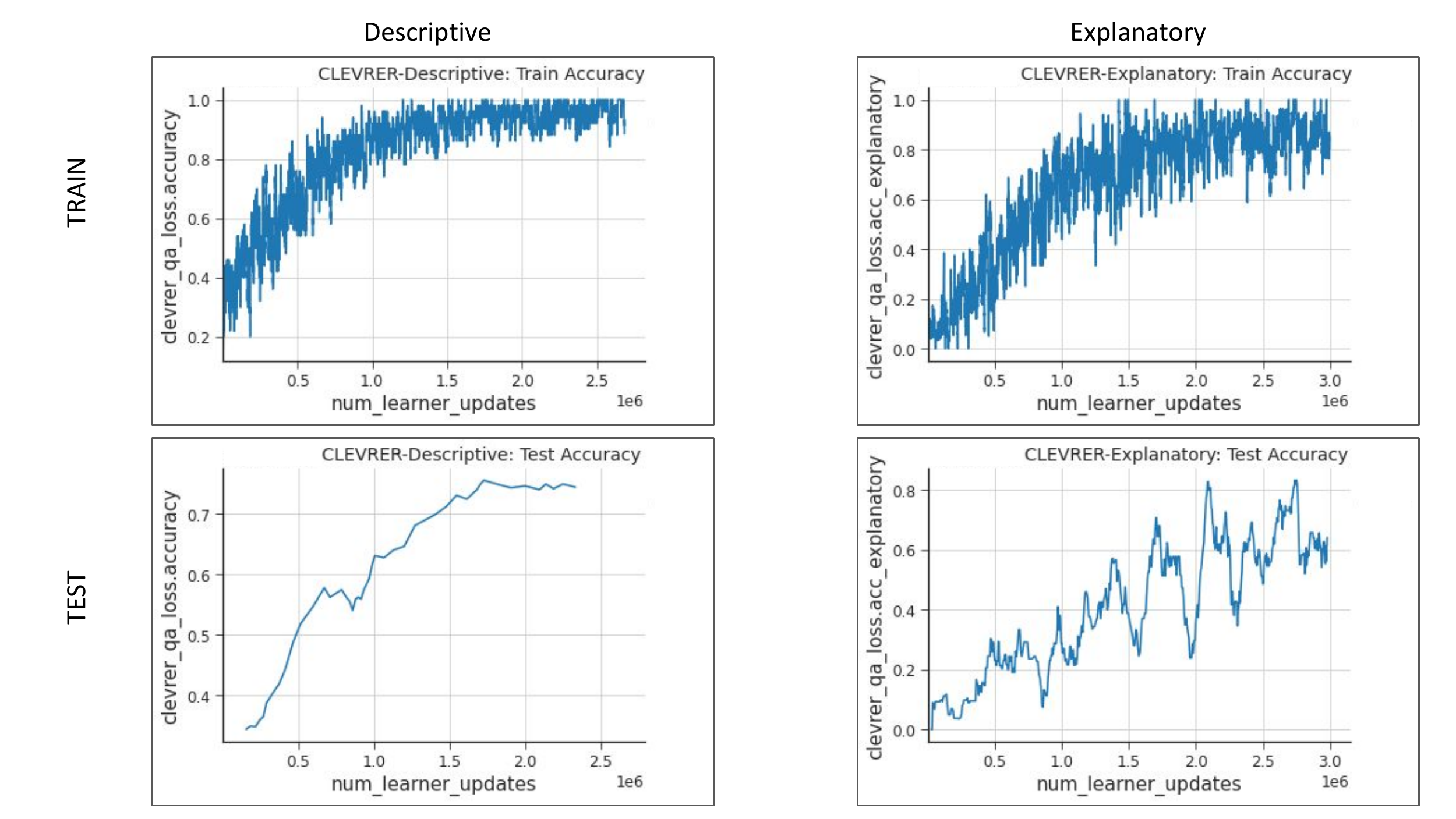}
\caption{Clevrer Train \& Evaluation Results on an $8$ slot, $4$ iteration model on \textit{Descriptive} and \textit{Explanatory} tasks.}
\label{fig:desc_exp}
\end{figure}

\begin{figure}[h!]
\centering
\includegraphics[width=1\textwidth]{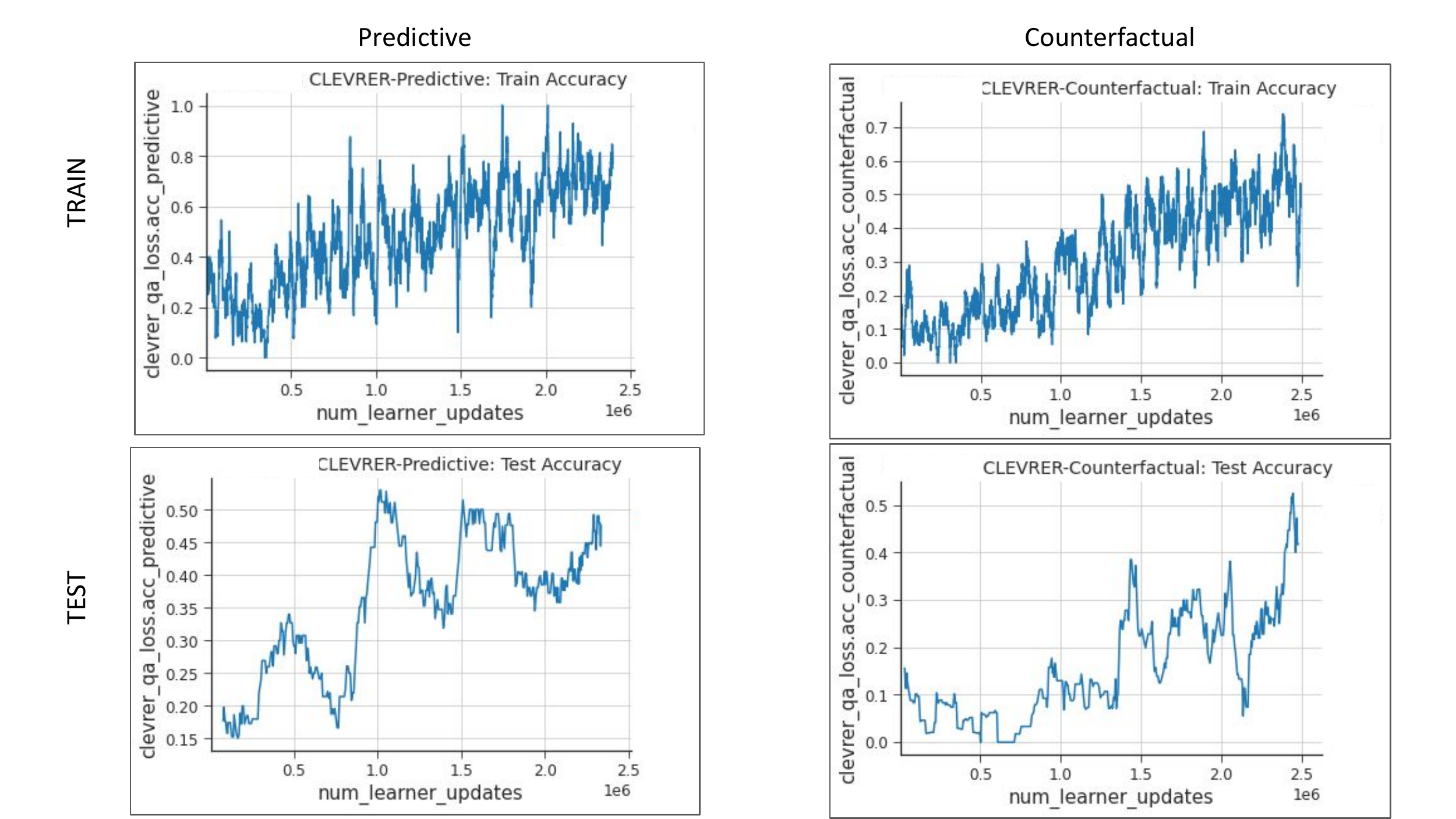}
\caption{Clevrer Train \& Evaluation Results on an $8$ slot, $4$ iteration model on \textit{Predictive} and \textit{Counterfactual} tasks.}
\label{fig:pred_cfctl}
\end{figure}

\begin{figure}[h!]
\centering
\includegraphics[width=1\textwidth]{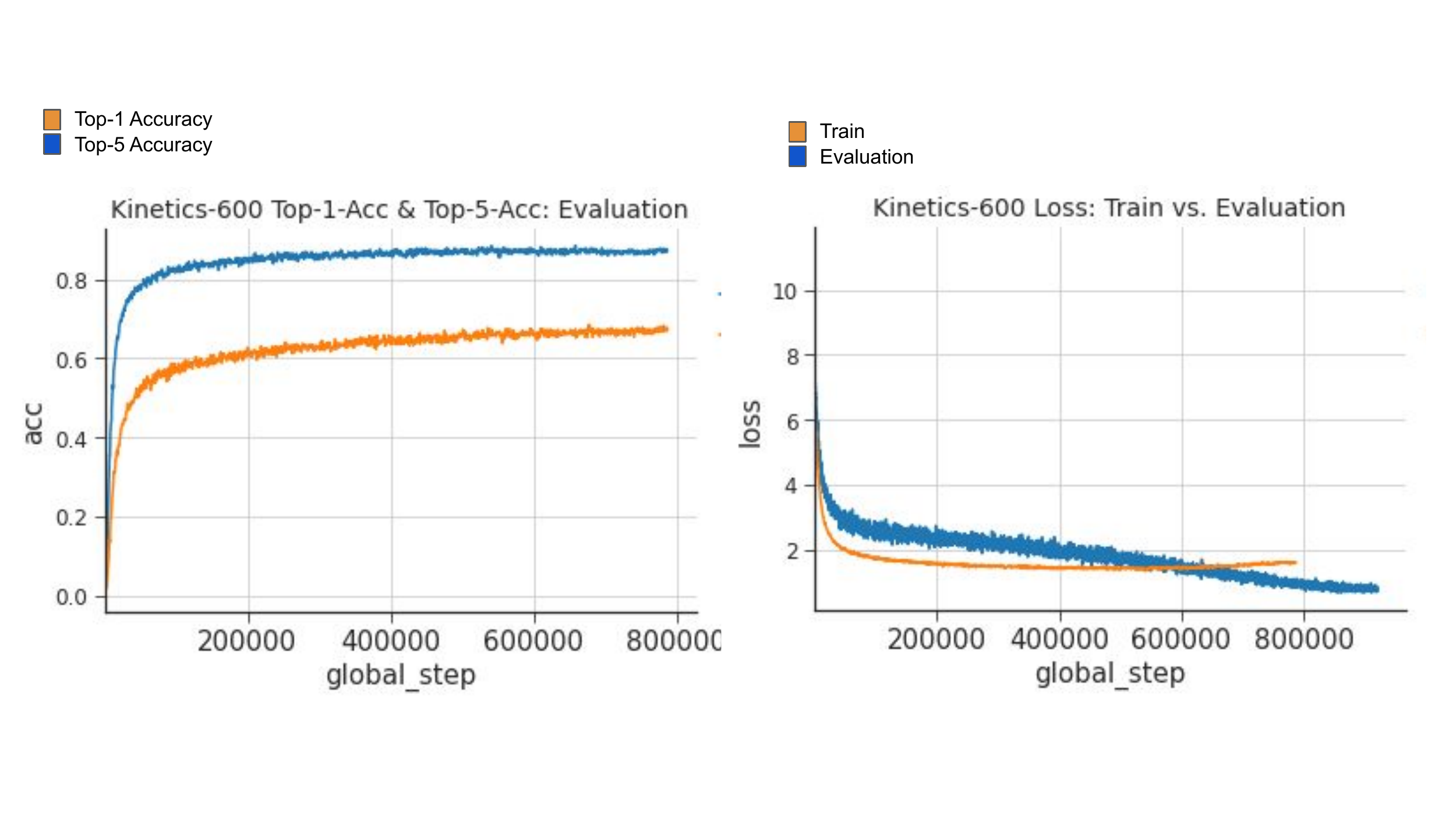}
\caption{Kinetics Top-5 \& Top-1 Accuracy and Train vs. Evaluation Losses.}
\label{fig:kinetics}
\end{figure}

We ran the analysis on the CATER static camera dataset.  In figures \ref{fig:lstm_masks} and \ref{fig:lstm_recon} we see the models produce qualitatively different masks and reconstructions.  There is notably much more blur and missing elements such as shadows in the LSTM model's reconstructions and the slot masks for the LSTM model also tend to be distributed among many objects in the scene.  This may indicate the representations learned in the LSTM model have been induced to model spurious correlations which may be expected in the absence of direct pathways between temporal events afforded by the transformer architecture.  Finally we note the performance of each model, we report scores for these models when the LSTM model performance began to top out.  For the LSTM model: \textit{top-5 accuracy} $60.7\%$, \textit{top-1 accuracy} $36.9\%$, \textit{L1 distance} $2.29$, and for the Transformer: \textit{top-5 accuracy} $71.9\%$, \textit{top-1 accuracy} $39.6\%$, \textit{L1 distance} $1.77$.  We believe that the advantage of the Transformer architecture would only grow greater as we larger sequences and more complex scenes are used.

\section{Baseline Details}

In Section 4 of the main paper a number of baselines are compared against and here we give a bit more detail on them.  For CLEVRER more detail can be found in \citet{yi2020clevrer} and \citet{chen2021grounding}, and in \citet{cater} for CATER.

First a note on CLEVRER, some of the baselines take advantage of additional ground truth data from video annotations and labels from the program executor. These can include ground truth information about the image content such as shapes, colours, positions, or motion data.  As we are primarily concerned with constructing a high quality sequence encoder that can infer these properties we did not include them in our training.  This notably puts our model at a disadvantage when comparing to neuro-symbolic approaches however, we believed it an important point to optimize for the cases where we may not be able to expect such ground truth labeling exists.  As a future proof of concept we are considering experimenting using of this data.

\textbf{Hierarchical Conditional Relational Networks (HCRN)} \citet{le2020hierarchical} introduce this model that imposes a hierarchy of CRN networks on clips that are built up into the full video.   CRNs take some number of input objects and a condition (query) and generate an output where relations among the input objects may be learned.  In HCRNs the output of lower \textit{clip} level CRNs is fed as input to the higher \textit{video} level CRNs to produce a final output that is decoded to an answer.

\textbf{Memory Attend Control Network (MAC V \& MAC V+)} \citet{hudson2018compositional} introduce the MAC network architecture adapted function on video input.  The network processes input via its \textit{input unit} to produce encodings from a question and video.  Next, the MAC unit processes input by iteratively attending to the output from the input unit. The \textit{control unit} recurrently attends to parts of the input, determining which parts to \textit{reason} about, while the \textit{memory stores} stores the result of step $i$.  Finally an \textit{output unit} combines the output from MAC cell iteration to produce a final output.  

The slot transformer exceeds the MAC network on \textit{Predictive}, \textit{Explanatory}, and \textit{Counterfactual} tasks in CLEVRER both with and without labelled data and HCRN on all CLEVRER tasks as well as CNN+MLP and CNN+LSTM stacks. 

\textbf{Localized, Compositional Video Question Answering (TVQA+)} \citet{lei2019tvqa} introduce the TVQA dataset and a multi-stream neural network that can ingest multimodal input for video QA.  Each stream effectively is composed of an embedding module (a CNN in the case of the video frames) followed by an LSTM.  Embedding streams include video, question, and answer data and are then \textit{fused} via elementwise products and concatenation before being softmaxed to produce answer scores.  This model also makes use of annotation data in CLEVRER.

\textbf{I3D \& R3D LSTM} Inflated 3D ResNet models. \citet{i3d17, cater} use pooling and filtering layers increased in size, or \textit{inflated}, so as to retain additional capacity to model 3D spatio-temporal input. They also pre-train on ImageNet.  The model utilizes two \textit{streams}: one feed-forward on RGB frames, and another for optical flow data from the sequence.  For CATER, a ResNet is used instead of a CNN and obtains strong results making use of optical flow and ImageNet pre-training (\citet{ILSVRC15}). Our model is competitive with theirs and in some cases exceeding their scores on the training videos alone.

\textbf{NS-DR \& DCL} We discuss neuro-symbolic logic models in the main paper in section 2. In particular two strong baselines \textit{Neuro-Symbolic Dynamic Reasoning} (\citet{yi2020clevrer}) and the \textit{Dynamic Context Learner} (DCL/ DCL (Oracle)) (\citet{chen2021grounding}).  As discussed above we haven't used ground truth data when training our models so our scores are difficult to compare with the top baseline however, despite this, the performance of the slot transformer remains competitive.

\end{document}